%% file: main.tex
\documentclass[11pt]{article}

\usepackage[preprint]{acl}

\usepackage{times}
\usepackage{latexsym}
\usepackage[T1]{fontenc}
\usepackage[utf8]{inputenc}
\usepackage{microtype}
\usepackage{graphicx}

\usepackage{amsmath}
\usepackage{amssymb}
\usepackage{booktabs}
\usepackage{multirow}
\usepackage{colortbl}
\usepackage{tcolorbox}

\title{VRPRM: Process Reward Modeling via Visual Reasoning}

\author{
  \textbf{Xinquan Chen} \\
  {\normalfont\small Shanghai AI Laboratory} \\
  \And
  \textbf{Chongying Yue} \\
  {\normalfont\small Shanghai AI Laboratory;} \\
  {\normalfont\small Zhejiang University} \\
  \And
  \textbf{Bangwei Liu} \\
  {\normalfont\small Shanghai AI Laboratory;} \\
  {\normalfont\small East China Normal University} \\
  \AND
  \textbf{Xuhong Wang}\thanks{\raggedright Corresponding author: \href{mailto:wangxuhong@pjlab.org.cn}{\texttt{wangxuhong@pjlab.org.cn}}} \\
  {\normalfont\small Shanghai AI Laboratory} \\
  \And
  \textbf{Yingchun Wang} \\
  {\normalfont\small Shanghai AI Laboratory} \\
  \And
  \textbf{Chaochao Lu} \\
  {\normalfont\small Shanghai AI Laboratory} \\
}

\begin{document}

\maketitle

% Unmarked footnote directly below the corresponding-author note:
% released code and the VRPRM model/data collections.
\begingroup
\renewcommand{\thefootnote}{}%
\footnotetext{\raggedright Code: \url{https://github.com/two-tiger/VRPRM}. VRPRM Collections: \url{https://huggingface.co/collections/two-tiger/vrprm}.}%
\endgroup

\begin{abstract}
% 中文：过程奖励模型（PRM）能够对中间推理步骤提供细粒度监督，但多数模型缺乏长程推理和深度思考能力。尽管已有工作为奖励模型引入思维链（CoT），高昂的 CoT-PRM 标注成本仍限制其扩展。为了应对上述挑战，我们提出视觉推理过程奖励模型 VRPRM，并设计了标注高效的两阶段训练策略。实验结果表明，仅使用 5.5K 个 CoT-PRM 监督步骤标签和 40K 条 non-CoT PRM 强化学习样本，VRPRM在 VisualProcessBench 上达到 69.18 macro F1，超越使用 400K 数据训练的非思考型 VisualPRM-8B 7.00 分。在 Best-of-N 评测中，VRPRM-RL 在 InternVL2.5-8B、26B 和 38B 策略上分别带来 21.77、23.42 和 23.12 分的 Overall 绝对提升，最高相对提升 59.0\%。
Process Reward Models (PRMs) provide fine-grained supervision over intermediate reasoning steps, yet most existing models lack long-range reasoning and deep thinking capabilities. Although prior work has introduced Chain-of-Thought (CoT) to reward models, the high annotation cost of CoT-PRM still limits its scalability. To address these challenges, we propose the Visual Reasoning Process Reward Model (VRPRM) and design a label-efficient two-stage training strategy. Experimental results demonstrate that with only 5.5K CoT-PRM supervised step labels and 40K non-CoT PRM reinforcement learning training examples, VRPRM achieves a macro F1 score of 69.18 on VisualProcessBench, outperforming VisualPRM-8B, a non-thinking PRM trained on 400K examples, by 7.00 points. In Best-of-N evaluation, VRPRM-RL brings absolute Overall improvements of 21.77, 23.42, and 23.12 points on InternVL2.5-8B, 26B, and 38B policies, respectively, with a maximum relative improvement of 59.0\%.
\end{abstract}

\section{Introduction}
% 中文：奖励模型（RM）是大语言模型（LLM）通过人类反馈强化学习（RLHF）进行后训练的核心组件。然而，现有奖励模型大多是仅评估最终结果的结果奖励模型（ORM）。这类模型只能为完整输出提供整体分数，难以监督生成过程中的关键推理步骤和内部逻辑结构，因而无法在强化学习期间为推理链质量提供稳定的奖励信号。
Reward Models (RMs) are a core component of Large Language Model (LLM) post-training through Reinforcement Learning from Human Feedback (RLHF). Most existing reward models, however, are Outcome Reward Models (ORMs) that evaluate only the final answer. A single holistic score cannot directly supervise critical intermediate steps or the internal logical structure of a generated solution, making it difficult to provide stable reward signals for reasoning quality during reinforcement learning.

% 中文：因此，越来越多的过程奖励模型（PRM）被提出，用于直接评价生成内容中的每一步。然而，一个关键问题仍未解决：缺乏推理能力的奖励模型如何可靠地指导具备思考能力的策略模型？
Process Reward Models (PRMs) address this limitation by assigning rewards to individual reasoning steps. Yet a central question remains: how can a reward model that does not itself reason reliably guide a reasoning-capable policy model?

% 中文：为提升奖励模型的能力与泛化性，现有工作分为两条互不相交的路径。思维链奖励模型在打分前引入显式推理，但仍属于结果级或偏好级评判：它们分析完整回答，却从不监督单个解题步骤。过程奖励模型则验证每一步，但依赖不含推理的标量打分，且均未采用强化学习训练。如表 1 所示，这两条路径尚未交汇。GenPRM 迈向推理型 PRM，但仅限文本、仅用 SFT，其合成的步骤级数据依赖强模型的高成本蒸馏——这一瓶颈限制了 CoT-PRM 监督的规模与任务多样性。VRPRM 填补这一空缺，让评判者自身先推理、从而可靠地指导具备推理能力的策略模型：少量质量受控的 CoT 监督激活全局推理，成本更低的过程级强化学习扩展步骤验证。
To improve reward-model capability and generalization, two lines of work have emerged. Chain-of-Thought reward models introduce explicit reasoning before scoring, yet remain outcome- or preference-level judges~\cite{guo2025reward,chen2025rm,hong2025think,zhang2025r1,wang2025bunified}: they analyze a complete response but never supervise individual steps. Process reward models~\cite{ong2025training,wang2025visualprm,wang2026athena}, in contrast, verify every step, but with scalar scores that involve no reasoning, and none is trained with RL. As summarized in Table~\ref{Table:compareRM}, these two lines have not intersected. GenPRM~\cite{zhao2025genprm} moves toward reasoning PRMs, yet is text-only and SFT-only, and its synthesized step-level data requires costly distillation---a bottleneck limiting the scale and diversity of CoT-PRM supervision. VRPRM occupies this open intersection, making the judge itself reason to reliably guide reasoning-capable policies: quality-controlled CoT supervision activates global reasoning, while lower-cost process-level RL scales step verification.

\begin{table}[t]
\centering
\scriptsize
\begin{tabular}{lcccc}
\toprule
\textbf{Reward Model} & \textbf{PRM} & \textbf{MM} & \textbf{CoT} & \textbf{RL} \\
\midrule
RRM~\cite{guo2025reward}                      &             &            & \checkmark & \checkmark \\
RM-R1~\cite{chen2025rm}                       &             &            & \checkmark & \checkmark \\
Think-RM~\cite{hong2025think}                 &             &            & \checkmark & \checkmark \\
R1-Reward~\cite{zhang2025r1}                  &             & \checkmark & \checkmark & \checkmark \\
UnifiedReward~\cite{wang2025bunified}         &             & \checkmark & \checkmark & \checkmark \\
Qwen-VL-PRM-7B~\cite{ong2025training}         & \checkmark  & \checkmark &            &            \\
GenPRM~\cite{zhao2025genprm}                  & \checkmark  &            & \checkmark &            \\
VisualPRM~\cite{wang2025visualprm}            & \checkmark  & \checkmark &            &            \\
Athena-PRM~\cite{wang2026athena}            & \checkmark  & \checkmark &            &            \\
\textbf{VRPRM (ours)}                                   & \checkmark  & \checkmark & \checkmark & \checkmark \\
\bottomrule
\end{tabular}
% 中文：不同奖励模型的能力比较。VRPRM 同时支持过程级评估、多模态输入、显式思维链和强化学习训练。
\caption{Comparison of representative reward models. VRPRM combines process-level evaluation, multimodal inputs, explicit CoT reasoning, and RL training. \textbf{PRM} denotes process-level evaluation; \textbf{MM}, multimodality; \textbf{CoT}, explicit reasoning; and \textbf{RL}, reinforcement-learning-based training.}
\label{Table:compareRM}
\end{table}

\begin{figure}[t]
    \centering
    \includegraphics[width=0.92\linewidth]{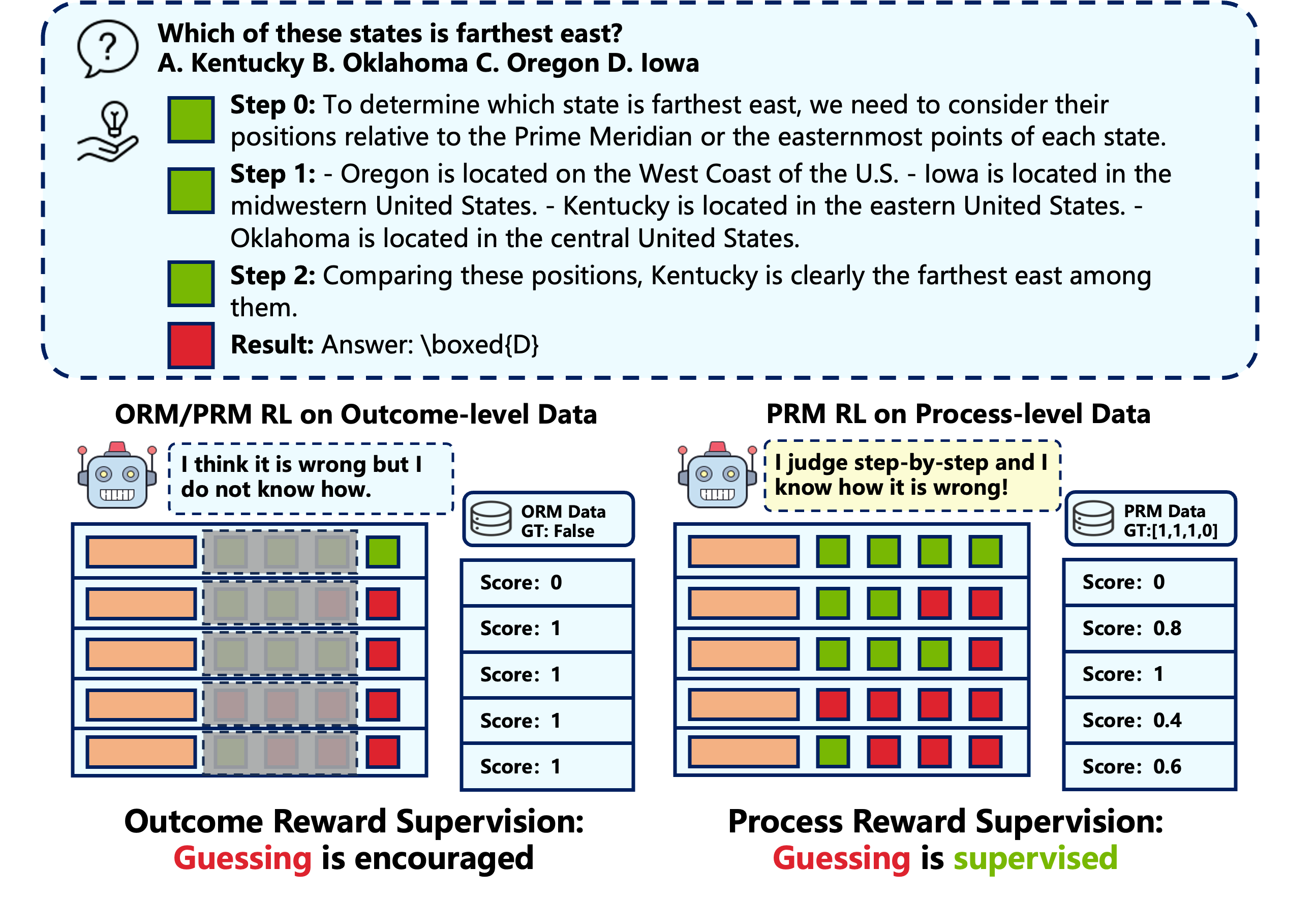}
    % 中文：为何需要过程级奖励。上：一条中间步骤合理但最终答案错误的轨迹。左：结果级监督只有一比特真值，评判者“知道它错了但不知道错在哪”，包括侥幸猜中在内的整条轨迹获得无差别奖励。右：步骤级真值支持分级奖励（0.4--1.0），区分部分一致与完全对齐的判断并定位首个错误；二元结果任务中 ORM 仍以 50\% 概率奖励随机猜测。
    \caption{\textbf{Why process-level rewards for RL.} Top: a trajectory with sound intermediate steps but a wrong final answer. Left: outcome supervision reduces to one bit of ground truth, so the judge ``knows it is wrong but not how,'' and whole trajectories---including lucky guesses---receive undifferentiated rewards. Right: step-level ground truth supports graded rewards (0.4--1.0) that separate partial agreement from exact judgment and pinpoint the first error; in a binary outcome task, an ORM still rewards a random guess with probability 50\%.}
    \label{fig:intro}
\end{figure}

% 中文：过程级监督也使强化学习天然适合 PRM 训练。如图 1 所示，结果奖励只提供一比特监督且可能奖励侥幸猜中的答案，而步骤级真值产生分级且与步骤对齐的奖励并定位首个错误。此外，这类过程标签可由蒙特卡洛 rollout 自动估计，因此强化学习避开了限制纯 SFT CoT-PRM 的高昂 CoT 标注成本。
Process-level supervision also makes RL a natural fit for PRM training. As Figure~\ref{fig:intro} illustrates, an outcome reward provides only one bit of supervision and can reward a lucky guess, whereas step-level ground truth yields graded, step-aligned rewards that pinpoint the first error. Moreover, such process labels can be estimated automatically by Monte Carlo rollouts, so RL avoids the costly CoT annotation that limits SFT-only CoT-PRMs~\cite{chen2025rm,chu2025sft}.

% 中文：本文提出视觉推理过程奖励模型 VRPRM，一个具备显式思维链的多模态 PRM，并采用标注高效的两阶段训练策略：少量质量受控的 CoT-PRM 对话激活全局推理，再在高置信的过程标签上进行强化学习以扩展步骤判别——总计仅使用 5.5K 个监督步骤标签和 40K 条非 CoT 样本。VRPRM 在 VisualProcessBench 上达到 69.18 的总体 macro F1；作为 Best-of-8 评价模型时，VRPRM-RL 在三种 InternVL2.5 策略规模上将 Overall 绝对提升 21.77--23.42 分。
We introduce Visual Reasoning PRM (VRPRM), a multimodal PRM with explicit CoT reasoning, trained with a label-efficient two-stage strategy: a small set of quality-controlled CoT-PRM conversations activates global reasoning, and RL on high-confidence process labels further scales step discrimination---in total, only 5.5K supervised step labels and 40K non-CoT examples. VRPRM reaches 69.18 overall macro F1 on VisualProcessBench; as a Best-of-8 critic, VRPRM-RL yields absolute Overall gains of 21.77--23.42 points across three InternVL2.5 policy scales.

% 中文：本文的主要贡献如下。
Our contributions can be summarized as follows:

\begin{itemize}
    % 中文：据我们所知，VRPRM 是首个通过强化学习训练的多模态 CoT-PRM，旨在提升奖励模型的细粒度推理与评估能力。
    \item \textbf{Pioneering the Integration of CoT RL in Visual PRMs.} To our knowledge, VRPRM is the first multimodal CoT-PRM trained with RL, explicitly designed to enhance the fine-grained reasoning and evaluation capabilities of reward models.

    % 中文：提出标注高效的两阶段训练策略。仅使用 5.5K 个监督步骤标签和 40K 条非 CoT 强化学习样本，VRPRM 便在 VisualProcessBench 上达到 69.18 macro F1，比使用 400K 样本训练的 VisualPRM-8B 高 7.00 分。
    \item \textbf{A Label-Efficient Two-Stage Training Strategy.} With only 5.5K supervised step labels and 40K non-CoT PRM examples for RL, VRPRM reaches 69.18 macro F1 on VisualProcessBench, exceeding VisualPRM-8B trained on 400K examples by 7.00 points.

    % 中文：提出有效的测试时扩展方法。VRPRM-RL 在六个多模态基准的 Best-of-8 评测中，将三种策略规模的 Overall 绝对提升 21.77--23.42 分，最高相对提升 59.0\%，展示了扩展模型能力的新途径。
    \item \textbf{An Effective Test-Time Scaling Approach.} VRPRM-RL improves Best-of-8 Overall by 21.77--23.42 absolute points across three policy scales, with a relative gain of up to 59.0\% over the base policy on six multimodal benchmarks.
\end{itemize}

\section{Related Work}
% 中文：过程奖励模型在强化学习优化和测试时扩展中日益重要。与只为最终输出分配单一分数的 ORM 不同，PRM 评估生成过程中的每个中间步骤，并将步骤级信号聚合为最终奖励，从而提供更细致的监督并减少“伪正确”问题。Qwen-Math-PRM 结合蒙特卡洛估计与大语言模型判断来筛选过程级标注数据；VisualPRM 使用 InternVL2.5 生成解题步骤，并通过蒙特卡洛采样评估步骤准确性。
\textbf{Process Reward Models.} Process Reward Models (PRMs) are increasingly important for reinforcement-learning optimization and test-time scaling. Unlike Outcome Reward Models (ORMs)~\cite{whitehouse2025j1,wang2025aunified,wang2025worldpm,zhang2024generative}, which assign one score to a final output, PRMs evaluate intermediate generation steps. Aggregating these step-level signals provides more informative supervision and mitigates ``spurious correctness,'' in which flawed reasoning reaches the correct answer. Qwen-Math-PRM~\cite{zhang2025lessons} combines Monte Carlo estimation with LLM judgments to construct process-level SFT data. VisualPRM~\cite{wang2025visualprm} uses InternVL2.5 models to generate solutions and Monte Carlo sampling to estimate step correctness.

% 中文：近期视觉语言 PRM 从不同角度提高视觉扎根和验证可靠性：Qwen-VL-PRM 将蒙特卡洛搜索与强视觉语言模型判断结合，TIM-PRM 引入外部工具和独立提问，Athena-PRM 利用强弱回答器之间的一致性筛选高置信过程标签，EVPV-PRM 则显式验证视觉前提。这些方法改善了监督质量，但大多仍采用标量或逐步验证范式；相比之下，VRPRM 最终生成一次可复用的全局视觉推理，再进行严格对齐的单词元步骤判断。
Recent vision-language PRMs improve visual grounding and verification reliability from complementary directions. Qwen-VL-PRM~\cite{ong2025training} combines Monte Carlo search with judgments from a strong VLM; TIM-PRM~\cite{kuang2025timprm} introduces external tools and independent evidence-seeking questions; Athena-PRM~\cite{wang2026athena} filters process labels using consistency between weak and strong completers; and EVPV-PRM~\cite{wang2026grounding} explicitly verifies visual premises. These methods improve supervision quality but largely retain scalar or step-local verification. In contrast, VRPRM generates reusable global visual reasoning once and then performs strictly aligned single-token step judgments.

% 中文：奖励建模研究正从传统标量打分转向生成推理链以辅助偏好判断的思维链奖励模型。RRM 将奖励建模视为推理任务，并通过强化学习提升推理能力。RM-R1 和 Think-RM 采用先监督微调、再强化学习的两阶段方案。后续工作将 CoT-ORM 扩展到多模态场景，例如 R1-Reward 和 UnifiedReward-Think。GenPRM 则在 PRM 中使用显式思维链和代码验证，但未采用强化学习。CoT 能提升奖励模型的可解释性和泛化性，但高质量 CoT 数据的获取与标注成本较高。
\noindent \textbf{Chain-of-Thought Reward Models.} Reward modeling has recently shifted from scalar scoring toward Chain-of-Thought Reward Models (CoT-RMs), which generate reasoning before making a preference judgment. RRM~\cite{guo2025reward} formulates reward modeling as a reasoning task and uses RL to strengthen that capability. RM-R1~\cite{chen2025rm} and Think-RM~\cite{hong2025think} follow a two-stage recipe: SFT on high-quality long-form reasoning data, followed by RL. R1-Reward~\cite{zhang2025r1} extends this recipe to multimodal data annotated by GPT-4o, while UnifiedReward-Think~\cite{wang2025bunified} combines multimodal preference data with RL. GenPRM~\cite{zhao2025genprm} introduces explicit CoT and code verification into a PRM but does not use RL. Although CoT improves interpretability and generalization, collecting high-quality reasoning traces remains expensive.

\section{Methodology}
\subsection{Problem Formulation}

% 中文：令 \(\mathcal{D}=\{(I,P,S)\}\) 表示由图像 \(I\)、问题 \(P\) 和解答 \(S\) 构成的数据集。每个解答由多个步骤组成，记为 \(S=(s_1,s_2,\dots,s_n)\)，其中 \(s_i\) 表示第 \(i\) 个步骤。
Let \(\mathcal{D}=\{(I,P,S)\}\) denote a dataset of images \(I\), problems \(P\), and candidate solutions \(S\). Each solution contains a sequence of steps, \(S=(s_1,s_2,\dots,s_n)\), where \(s_i\) is the \(i\)-th step.

% 中文：VisualPRM 将过程评估建模为多轮对话，以利用多模态大语言模型的生成能力，并将标签词元“1”的预测概率作为步骤得分。令 \(M\) 表示视觉 PRM，则第 \(i\) 步得分可形式化如下。
\noindent \textbf{Visual PRM.} VisualPRM~\cite{wang2025visualprm} formulates process evaluation as a multi-turn dialogue and uses the probability assigned to label token ``1'' as the score of each step. Let \(M\) denote a visual PRM. The score for step \(i\) is
\begin{equation}
    y_i = P_M(1 \mid I,P,s_{\leq i}),
\end{equation}
% 中文：其中，\(y_i\) 表示第 \(i\) 步的得分；通过设定阈值即可判断该步骤是否正确。
where \(y_i\) is the score of the \(i\)-th step. A threshold converts this score into a binary correctness judgment.

% 中文：视觉推理 PRM 在视觉 PRM 基础上加入显式思维链。给定图像 \(I\)、问题 \(P\) 和解答 \(S\)，模型首先生成推理轨迹 \(\mathcal{T}\)，随后输出结构化评估结果 \(\mathcal{R}\)。
\noindent \textbf{Visual Reasoning PRM.} We extend a visual PRM with an explicit reasoning trace, such as CoT~\cite{wei2022chain}. Given \(I\), \(P\), and \(S\), VRPRM first generates a reasoning trace \(\mathcal{T}\) and then produces a structured evaluation \(\mathcal{R}\):
\begin{equation}
    \mathcal{T} \sim \pi_{\theta}(I,P,S), \qquad
    \mathcal{R} \sim \pi_{\theta}(I,P,S,\mathcal{T}).
\end{equation}
% 中文：从结构化输出 \(\mathcal{R}\) 中解析得到步骤级过程奖励 \((r_1,\dots,r_n)\)。
We parse \(\mathcal{R}\) to obtain the step-level rewards \((r_1,\dots,r_n)\).

\begin{figure*}[t]
    \centering
    \includegraphics[width=\textwidth]{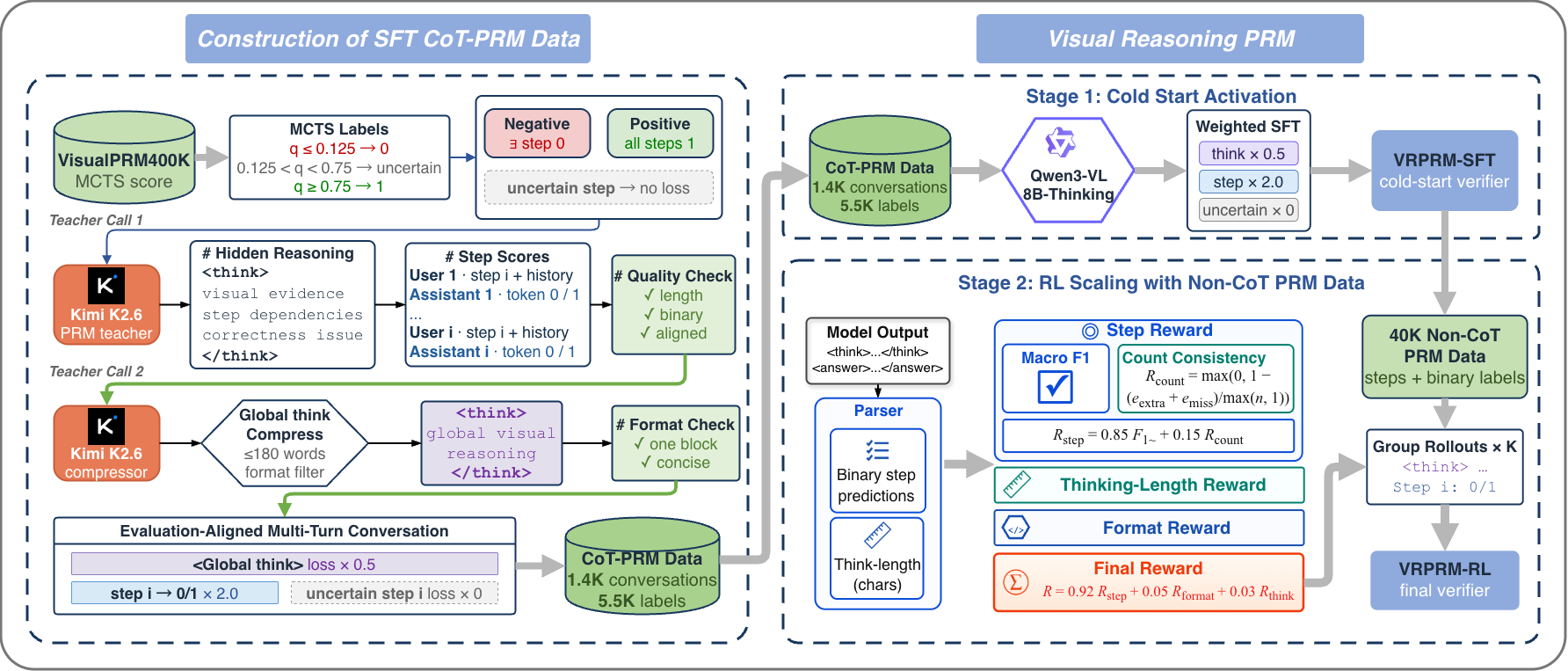}
    % 中文：VRPRM 总体框架。左侧使用双阈值离散化 VisualPRM400K 的 MCTS 分数，并通过 Kimi-K2.6 的教师判断与思考压缩两次调用构造评测对齐的多轮 CoT-PRM 数据；高置信步骤采用源标签，不确定步骤保留为连续历史但屏蔽损失。右侧首先以 0.5/2.0/0 的权重对全局思考、高置信步骤和不确定步骤进行 SFT，得到 VRPRM-SFT；随后在 40K 条高置信非 CoT 样本上进行组 rollout，并结合以 macro F1 和数量一致性为主的步骤奖励、格式奖励及思考长度奖励，得到 VRPRM-RL。
    \caption{\textbf{Overview of VRPRM.} Left: dual-thresholded MCTS labels and two Kimi-K2.6 calls produce evaluation-aligned multi-turn CoT-PRM data; uncertain judgments remain as loss-masked history. Right: weighted SFT yields VRPRM-SFT, followed by group-based RL on 40K high-confidence non-CoT examples with structured step, format, and thinking-length rewards.}
    \label{fig:arch}
\end{figure*}

\subsection{Cold-Start Reasoning Activation}
% 中文：指令微调后的 LLM 虽能通过提示完成基本过程评估，但一次性生成完整推理和全部步骤得分时容易出现格式错误或步骤数量不一致。因此我们将任务分解为一次全局思考和一系列单词元步骤判断，并在冷启动训练与评测中采用相同的轮次结构和单词元判断协议。
Instruction-tuned LLMs can perform basic process evaluation from prompts, but jointly generating a full rationale and all step scores often causes malformed outputs or a mismatch between the number of judgments and solution steps. We instead factorize evaluation into one global-thinking turn followed by a sequence of single-token step judgments, using the same turn structure and single-token decision protocol for cold-start training and evaluation.

\subsubsection{Synthetic CoT-PRM Data}

% 中文：我们以 VisualPRM400K-v1.1-Raw 构建 SFT 数据。每个源样本包含一张或多张图像、问题、参考答案、候选解答步骤以及 MCTS 得到的步骤分数。由于这些分数并非人工标注的二值标签，直接将所有分数用于监督会引入大量模糊性，因此针对正负例设置阈值进行过滤：
We construct the SFT data from VisualPRM400K-v1.1-Raw~\cite{wang2025visualprm}. Each source example contains one or more images, a question, a reference answer, a candidate solution with step-wise reasoning, and MCTS-derived step scores \(q_i\). Since these scores are not human-annotated binary labels, directly using all of them as supervision would introduce substantial ambiguity. We therefore filter the steps with positive and negative thresholds:
\begin{equation}
\tilde{y}_i =
\begin{cases}
0, & q_i \leq \tau_-=0.125,\\
1, & q_i \geq \tau_+=0.75,\\
\varnothing, & 0.125 < q_i < 0.75,
\end{cases}
\end{equation}
% 中文：其中空标签表示不确定步骤，不进入监督目标。至少包含一个负步骤的样本进入负源池；所有步骤均为正的样本进入正源池。正源池提供完整正确过程，避免模型过度预测 0。
where \(\varnothing\) denotes an uncertain step excluded from the supervised objective. We form a negative-source pool from examples containing at least one negative step and a positive-source pool from examples whose steps are all positive. The positive pool supplies complete correct trajectories and prevents the model from over-predicting 0.

% 中文：针对每个选定的源示例，我们对 Kimi-K2.6 进行两次查询。在首次调用中，Kimi 充当视觉 PRM 教师的角色，综合考察图像、问题、参考答案以及完整的候选解法。其隐式推理过程会分析相关的视觉证据、答案约束条件、步骤间的依赖关系、首个导致后续错误的偏差，以及后续步骤是独立有效还是承袭了先前的错误。随后 Kimi 模型根据每一步解法和此前的隐式推理，输出严格的 0/1 二值判断。
For each selected source example, we query Kimi-K2.6~\cite{kimi2025k2} twice. In the first call, Kimi acts as a visual PRM teacher and jointly examines the images, the question, the reference answer, and the complete candidate solution. Its hidden reasoning analyzes the relevant visual evidence, the answer constraints, dependencies among steps, the first deviation that causes subsequent errors, and whether later steps are independently valid or inherit an earlier mistake. Kimi then outputs a strict binary \{0,1\} judgment for every step based on the solution and its prior implicit reasoning.

% 中文：仅当回复可解析、分数数量与候选步骤严格一致、每个分数均为 0/1，且与 MC 标签一致时才接受；否则拒绝并重试，耗尽重试次数的样本记入失败集合并排除。因此 Kimi 不重新标注高置信步骤，最终训练标签始终由保守 MC 标签控制，其作用是在标签一致的前提下提供全局推理。
A response is accepted only if it is parseable, contains exactly one score per candidate step, has binary \{0,1\} values only, and agrees with the MC labels; otherwise it is rejected and retried, and examples that exhaust the retry budget are recorded as failures and excluded. Thus, Kimi does not relabel the confident steps: the final training labels remain governed by the conservative MC labels, and the teacher's role is to supply global reasoning consistent with them.

% 中文：第二次调用 Kimi 将第一次过长的隐藏推理压缩为一个简洁、可见的全局思考块。这种浓缩处理保留了关键的视觉证据、参考答案的约束条件、步骤间的依赖关系和主要错误点，同时将解题思路的篇幅限制在 100 到 2,000 个字符之间。此过程确立了一个明确的全局思考目标，旨在引导学生在对具体步骤做出判断之前，先分析完整的解题方案。数据集统计信息及具体的提示词（prompt）详见附录~\ref{Appendix:RollPrompt}。
The second call compresses the overly long hidden reasoning from the first call into a concise, visible global-thinking block. This compression preserves the key visual evidence, the reference-answer constraints, dependencies among steps, and the principal errors, while limiting the rationale to between 100 and 2,000 characters. It establishes an explicit global-thinking target that guides the student to analyze the complete solution before judging individual steps. Dataset statistics and the exact prompts are given in Appendix~\ref{Appendix:RollPrompt}.

\subsubsection{Supervised Fine Tuning}

% 中文：每个通过筛选的源样本最终合并为一条与评测对齐的多轮对话。首轮用户输入包含图像、问题、参考答案和完整候选解答，助手生成一次全局思考；之后每轮用户输入当前步骤与此前判断，助手只输出单词元 0 或 1。该分解避免一次性生成推理和全部标签时的格式与数量错误。
Each filtered source example is finally merged into a single evaluation-aligned multi-turn conversation. The initial user turn provides the images, the question, the reference answer, and the complete candidate solution, and the assistant responds with one global-thinking block. Each subsequent user turn supplies the current step together with all previous judgments, and the assistant outputs exactly one token, 0 or 1. This factorization avoids the formatting and step-count errors that arise when the rationale and all labels are generated at once.

% 中文：高置信步骤使用源 MC 二值标签。不确定步骤仍以 Kimi 的二值判断保留在对话中，使后续步骤能够读取完整、连续的历史，但其助手回复不参与损失。因此 Kimi 对不确定位置的预测只作为上下文状态，绝不作为硬监督。
Confident steps use the source Monte Carlo binary labels. Uncertain steps are still retained in the conversation with Kimi's binary judgments so that subsequent steps read a complete, uninterrupted history, but their assistant responses are excluded from the loss. Kimi's predictions at uncertain positions therefore serve only as contextual state and never as hard supervision.

% 中文：我们使用上述多轮数据初始化长程推理和过程评估能力。为避免较长的全局思考文本主导优化，同时强化真正决定过程指标的步骤标签，对两类助手目标采用不同权重。
We use these multi-turn data to initialize long-horizon reasoning and process evaluation. To prevent the longer global-thinking text from dominating optimization while emphasizing the step labels that truly determine process metrics, the two assistant targets are weighted differently:
\begin{equation}
    \mathcal{L}_{\mathrm{SFT}}
    =0.5\!\sum_{j\in\mathcal{T}}\!\operatorname{CE}(z_j)
    +2.0\!\sum_{i\in\mathcal{C}}\!\operatorname{CE}(\tilde{y}_i),
\end{equation}
% 中文：其中，T 表示全局思考词元，C 表示具有高置信标签的步骤集合。系统、用户和图像词元均不计算损失，不确定步骤的助手回复同样被屏蔽。附录 A 与 C 给出详细提示词、序列化格式、数据统计和完整训练配置。
where \(\mathcal{T}\) denotes the global-thinking tokens and \(\mathcal{C}\) the set of steps with confident labels. System, user, and image tokens receive no loss, and the assistant responses of uncertain steps are likewise masked. Appendices~\ref{Appendix:RollPrompt} and~\ref{Appendix:ImplementationDetails} provide the detailed prompts, serialization format, data statistics, and complete training configuration.

% 中文：SFT 的作用是让模型习得“先全局审题、再逐步判分”的评测协议；但其监督目标是模仿第三方评分器产生的启发式标签，无法直接优化判分质量本身，这促使第二阶段的强化学习。
SFT instills the evaluation protocol---reasoning globally before judging each step---but its supervision imitates heuristic labels produced by a third-party scorer and cannot directly optimize scoring quality itself, which motivates the second-stage reinforcement learning.

\subsection{RL Scaling with Non-CoT PRM Data}
% 中文：RL 的数据同样取自 VisualPRM400K-v1.1-Raw，但 RL 训练不引入推理标注。我们仍先用保守阈值过滤 MC 分数：任一步骤落在中间区间 (0.125<q_i<0.75) 的整条样本被丢弃；随后在高置信的数据池中选择 40K 条训练样本，每条包含评分 prompt、images 和结构化 ground truth（构造细节见附录 B）。
The RL data is likewise derived from VisualPRM400K-v1.1-Raw~\cite{wang2025visualprm}, but RL training introduces no reasoning annotation. We still filter the Monte Carlo scores with the conservative thresholds: an entire example is discarded if any step falls in the intermediate zone (\(0.125<q_i<0.75\)). We then select 40K training examples from this high-confidence pool; each contains a scoring prompt, the images, and structured ground truth (construction details in Appendix~\ref{Appendix:RLReward}).

% 中文：prompt 同时提供问题和带原始索引的完整候选步骤。模型先生成一个简洁的全局 <think> 块，再在 <answer> 块中仅按 “Step i: 0/1” 的格式输出全部步骤标签。该结构保留 SFT 阶段学到的“全局思考后逐步判断”模式，同时使 rollout 可被确定性解析，并降低 RL 中的格式漂移。
The prompt provides the question together with the complete candidate steps and their original indices. The model first generates one concise global \texttt{<think>} block and then emits all step labels in an \texttt{<answer>} block using only the grammar ``Step \(i\): 0/1.'' This structure preserves the global-thinking-then-step-judgment pattern learned during SFT, makes rollouts deterministically parseable, and reduces format drift during RL.

% 中文：以 LoRA 合并后的 π_SFT 为初始策略，在同一任务上用组相对策略优化继续训练：对每个提示 (I,P,S) 采样 16 条 rollout，每条一次性生成全局推理与全部步骤判决 (T, r̂)，并由规则奖励函数评分。
Starting from the LoRA-merged \(\pi_{\mathrm{SFT}}\), we continue training on the same task with group-relative policy optimization~\cite{shao2024deepseekmath}: for each prompt \((I,P,S)\), 16 rollouts are sampled, each generating the global reasoning and all step judgments \((\mathcal{T},\hat{r})\) in one self-contained pass, and every rollout is scored by a rule-based reward:
\begin{equation}
    R(\tau)=w_s R_{\mathrm{step}}(\hat{r},\hat{y})+w_f R_{\mathrm{fmt}}+w_t R_{\mathrm{think}},
\end{equation}
% 中文：奖励由主导的步骤判分项与两个轻量辅助项（格式、思考长度）加权组成，权重 w=(0.92, 0.05, 0.03)。
which combines a dominant step-scoring term with two lightweight auxiliary terms, with weights \(w=(0.92,0.05,0.03)\).

% 中文：步骤奖励。核心项在每条源样本内部，将预测标签 r̂ 与过滤得到的确认真值标签 ŷ 对照打分：
\noindent\textbf{Step reward.} The core term scores the predicted labels \(\hat{r}\) against the confirmed ground-truth labels \(\hat{y}\) within each source example:
\begin{equation}
    R_{\mathrm{step}}=\lambda\min\!\big(\operatorname{F1}_{\mathrm{macro}}\!(\hat{r},\hat{y}),R_{\mathrm{cnt}}\big)+(1{-}\lambda)R_{\mathrm{cnt}}
\end{equation}
% 中文：macro F1 在每条样本内部的正、负两类上计算，防止模型坍缩到多数类；λ=0.85。
where the within-example macro F1 over the positive and negative classes prevents majority-class collapse, and \(\lambda=0.85\).

% 中文：数量一致性。e_extra 与 e_miss 分别表示多余与缺失的步骤索引：
\noindent\textbf{Count consistency.} With \(e_{\mathrm{extra}}\) and \(e_{\mathrm{miss}}\) denoting spurious and missing step indices,
\begin{equation}
    R_{\mathrm{cnt}}=\max\!\Big(0,\,1-\frac{e_{\mathrm{extra}}+e_{\mathrm{miss}}}{\max(n,1)}\Big),
\end{equation}
% 中文：对称惩罚两类错位，并在 R_step 内截断 macro F1——只输出少量容易步骤无法获得虚高奖励。
which symmetrically penalizes both types of misalignment and caps macro F1 inside \(R_{\mathrm{step}}\), so emitting only a few easy labels cannot earn a high reward.

% 中文：格式与思考长度。R_fmt 对完整 <think>/<answer> 结构给满分、对不完整结构给分档部分分；R_think 奖励 80--1,200 字符的适中推理而非单调奖励更长文本；分段定义见附录 B。
\noindent\textbf{Format and length.} \(R_{\mathrm{fmt}}\) gives full credit to a complete \texttt{<think>}/\texttt{<answer>} structure and graded partial credit to incomplete ones, while \(R_{\mathrm{think}}\) favors a moderate rationale of 80--1,200 characters rather than monotonically rewarding length; their piecewise definitions are given in Appendix~\ref{Appendix:RLReward}.

% 中文：优势为组内奖励标准化。
Advantages are estimated by within-group normalization,
\begin{equation}
    \widehat A_j=\frac{R(G,O_j)-\operatorname{mean}_{k}R(G,O_k)}
    {\operatorname{std}_{k}R(G,O_k)+\epsilon},
\end{equation}
% 中文：策略更新采用 token 级裁剪代理目标，并附加 KL 惩罚将策略锚定在 π_SFT 上。
and the policy is optimized with a clipped surrogate anchored to \(\pi_{\mathrm{SFT}}\) by a KL penalty,
\begin{equation}
\begin{aligned}
    \mathcal{J}={}&\mathbb{E}\Big[\sum_i \min\!\big(\rho_i \widehat A_i,\operatorname{clip}(\rho_i,1{-}\epsilon,1{+}\epsilon)\widehat A_i\big)\Big]\\
    &-\beta\,\mathrm{KL}\big(\pi_{\theta}\,\|\,\pi_{\mathrm{SFT}}\big),
\end{aligned}
\end{equation}
% 中文：其中 ρ_i 为由长度归一化序列似然比得到的 token 级重要性权重。
where \(\rho_i\) are token-level importance ratios derived from the length-normalized sequence likelihood~\cite{zheng2025gspo}.
% 中文：RL 阶段在 40K 提示池上运行 600 个优化步（每步 32 提示 × 16 rollout），最终 checkpoint 的选择见附录 C、D。与 SFT 的模仿学习不同，RL 直接以任务指标（macro F1）为优化目标，在保持协议不变的前提下提升判分质量；超参数见附录 C。
The RL stage runs 600 optimization steps on this 40K prompt pool (32 prompts $\times$ 16 rollouts per step). Unlike the imitation objective of SFT, RL directly optimizes a within-example macro-F1 reward over step judgments while preserving the learned protocol; hyperparameters are given in Appendix~\ref{Appendix:ImplementationDetails}.

\subsection{Test-Time Scaling}
% 中文：我们沿用 VisualPRM 的 Best-of-N 设置，使用 PRM 作为评价模型，从多个候选回答中选择最优回答。
We follow the Best-of-N (BoN) setup of VisualPRM~\cite{wang2025visualprm}, using a PRM as the critic that selects one response from multiple candidates.

% 中文：推理阶段，PRM 首先基于图像、问题和完整候选解答生成一次全局思考，随后复用该上下文逐步判断。每次只输入当前步骤和已有历史判断，并将输出约束为单词元 0 或 1；词元“1”的概率作为步骤奖励。
At inference time, the PRM first generates global thinking from the image, problem, and complete candidate solution. It then reuses this context for stepwise scoring, conditioning each turn on the current step and previous judgments while constraining generation to the single token 0 or 1. For \(S=(s_1,s_2,\dots,s_n)\), the reward at position \(t\) is
\begin{equation}
    r_t=P_{\theta}(1\mid I,P,S,\mathcal{T},\hat{y}_{<t},s_t).
\end{equation}
% 中文：其中，\(\mathcal{T}\) 为一次生成并复用的全局思考，\(\hat{y}_{<t}\) 为此前步骤判断。这一分解减少了重复推理开销，并从结构上保证每个请求只产生一个步骤判断。对 \(N\) 个候选回答分别计算平均过程得分并选择最高者。
Here, \(\mathcal{T}\) is generated once and reused, and \(\hat{y}_{<t}\) contains previous step judgments. This factorization reduces repeated reasoning and structurally guarantees one decision per requested step. For \(N\) candidates \(\{S_1,S_2,\dots,S_N\}\), we compute the mean process score of each and select
\begin{equation}
    S^{*}=\operatorname*{arg\,max}_{S_i \in \{S_1,\dots,S_N\}}
    \frac{1}{n_i}\sum_{t=1}^{n_i}r_t^{,i}.
\end{equation}

\begin{table*}[ht]
\centering
\scriptsize
\setlength{\tabcolsep}{4.2pt}
\begin{tabular}{@{}lccccccc@{}}
\toprule
\textbf{Model} & \textbf{\# Samples} & \textbf{DynaMath} & \textbf{MMMU} & \textbf{MathVerse} & \textbf{MathVision} & \textbf{WeMath} & \textbf{Overall} \\
\midrule
\multicolumn{8}{c}{\textbf{Proprietary Models}} \\
\midrule
GPT-4o-mini & -- & 56.57 & 54.08 & 52.53 & 51.42 & 56.74 & 53.57 \\
GPT-4o & -- & 56.30 & 60.20 & 59.70 & 59.00 & 63.30 & 60.30 \\
Doubao-Seed-1.6-Vision & -- & 66.19 & 59.47 & 63.12 & 61.07 & 62.74 & 62.77 \\
Gemini-2.0-Flash & -- & 58.50 & 60.10 & 62.80 & 66.70 & 58.70 & 62.30 \\
Gemini-2.5-Pro & -- & 68.47 & 63.34 & \underline{68.26} & 65.15 & 69.48 & 67.13 \\
\midrule
\multicolumn{8}{c}{\textbf{Open-source Models}} \\
\midrule
Qwen2.5-VL-72B-Instruct & -- & 56.99 & 59.43 & 56.43 & 58.09 & 55.72 & 57.19 \\
Qwen3-VL-30B-A3B-Instruct & -- & 58.95 & 61.29 & 57.37 & 57.49 & 58.76 & 58.22 \\
Qwen3-VL-235B-A22B-Instruct & -- & 57.63 & 58.73 & 58.08 & 59.59 & 58.76 & 58.51 \\
Qwen3-VL-8B-Thinking & -- & 44.89 & 43.25 & 40.76 & 41.19 & 45.85 & 42.48 \\
\midrule
\multicolumn{8}{c}{\textbf{Process Reward Models}} \\
\midrule
Qwen-VL-PRM-7B~\cite{ong2025training} & 290K & 58.30 & 55.80 & 58.80 & 55.70 & 59.80 & 58.60 \\
VisualPRM-8B~\cite{wang2025visualprm} & 400K & 61.57 & 59.57 & 61.71 & 62.81 & 60.69 & 62.18 \\
Athena-PRM~\cite{wang2026athena} & 155K & \textbf{72.70} & \textbf{74.10} & 60.70 & 61.20 & \textbf{73.80} & 65.90 \\
EVPV-PRM~\cite{wang2026grounding} & -- & 69.57 & \underline{68.86} & 67.09 & 65.27 & 69.11 & \underline{67.46} \\
\midrule
\textbf{VRPRM-SFT} & 5.5K & 65.19 & 60.94 & 66.91 & \underline{68.05} & 66.03 & 66.77 \\
\quad w/o Thinking & 5.5K & 64.02 & 56.26 & 61.14 & 61.25 & 61.77 & 61.63 \\
\quad + Cont.\ SFT & 5.5K+40K & 60.07 & 56.77 & 63.80 & 57.79 & 60.63 & 61.22 \\
\textbf{VRPRM-RL} & 5.5K+40K & \underline{72.31} & 60.12 & \textbf{68.32} & \textbf{69.01} & \underline{71.12} & \textbf{69.18} \\
\quad w/o Thinking & 5.5K+40K & 63.90 & 59.65 & 65.61 & 66.76 & 64.73 & 65.47 \\
\bottomrule
\end{tabular}
% 中文：VisualProcessBench 各子集上的 macro F1。Qwen3-VL-8B-Thinking 与各 thinking 变体使用相同的逐步单词元推理接口；w/o Thinking 在单次请求中对全部步骤打分，视觉输入相同；+ Cont. SFT 表示从 VRPRM-SFT 出发在同一 40K 池上继续监督训练。粗体和下划线分别表示每列的最优和次优结果。5.5K 表示来自 1,378 条多轮对话的可训练 SFT 步骤标签数，40K 表示第二阶段非 CoT 池；”--”表示训练样本数不适用或未报告。
\caption{\textbf{Macro F1 on VisualProcessBench.} Qwen3-VL-8B-Thinking and the thinking variants share the stepwise single-token inference interface; ``w/o Thinking'' scores all steps in one pass with the same visual input; ``+ Cont.\ SFT'' continues supervised training from VRPRM-SFT on the same 40K pool. \textbf{Bold} and \underline{underlined} values denote the best and second-best result in each column. The 5.5K count refers to trainable SFT step labels from 1,378 multi-turn conversations, and 40K denotes the second-stage non-CoT pool; ``--'' indicates not applicable or not reported.}
\label{Table:PRMBench}
\end{table*}

\section{Experiments}

\subsection{Experiment Settings}

% 中文：基础模型与训练流程。VRPRM 使用 Qwen3-VL-8B-Thinking 作为基础模型；选择该模型是为了利用其强大的多模态推理能力，为图像条件下的全局思考与逐步过程验证提供良好初始化。我们使用ms-swift框架进行监督微调，冻结了基础模型的视觉编码器，仅训练语言编码器的LoRA参数。再从所得冷启动检查点出发，使用 verl 框架进行强化学习，得到最终的 VRPRM-RL 检查点；RL 采用 AdamW 优化器和 KL 惩罚。
\noindent \textbf{Base Model and Training Pipeline.}
We employ Qwen3-VL-8B-Thinking~\cite{qwen2025qwen3vl} as the base model for VRPRM, selected for its powerful multimodal reasoning capabilities, which provide a favorable initialization for global thinking under image conditions and stepwise process verification. Supervised fine-tuning is conducted using the ms-swift framework~\cite{zhao2024swiftascalablelightweightinfrastructure}, where the visual encoder of the base model is frozen and only the LoRA parameters of the language encoder are trained. Starting from the resulting cold-start checkpoint, we perform reinforcement learning with the verl framework~\cite{sheng2024hybridflow} to obtain the final VRPRM-RL checkpoint; the RL phase adopts the AdamW optimizer with a KL term to the frozen reference.

% 中文：基准与指标。五个子集上报告 macro F1；Overall 列先池化五个子集的全部步骤，再按正/负两类计算 macro F1，因此大子集权重更高，并非子集列的均值；评测不向模型提供参考答案；另报告 Best-of-N 选择与词元/延迟开销；度量约定、各来源样本数与 VisualPRM-8B 复测细节见附录 C。
\noindent \textbf{Benchmarks and Metrics.}
We report macro F1 for step-correctness classification on five VisualProcessBench~\cite{wang2025visualprm} subsets---DynaMath~\cite{zou2025dynamath}, MMMU~\cite{yue2024mmmu}, MathVerse~\cite{zhang2024mathverse}, MathVision~\cite{wang2024measuring}, and WeMath~\cite{qiao2024we}. The Overall column pools all steps of the five subsets and then computes macro F1 over the correct and incorrect classes, so larger subsets carry proportionally more weight; it is not the mean of the per-subset columns. The reference answer is never shown to the model. We also report Best-of-N selection performance and token/latency overhead; metric conventions, per-source counts, and the VisualPRM-8B re-evaluation are detailed in Appendix~\ref{Appendix:ImplementationDetails}.

\subsection{VisualProcessBench Results}

% 中文：性能分析。表 2 在相同推理接口下将基础模型与两阶段训练结果直接对照。未经过程监督的基础模型仅取得 42.48 的总体 macro F1，说明全局思考与单词元输出格式本身并不能产生可靠的过程验证能力。仅使用 5.5K 个监督步骤标签，VRPRM-SFT 即达到 66.77，与此前最强 PRM 相差不足 1 分；RL 阶段将 VRPRM-RL 提升至 69.18，超过 EVPV-PRM 1.72 分、超过使用 400K 样本训练的 VisualPRM-8B 7.00 分。增益随子集变化：VRPRM-RL 在 MathVerse 与 MathVision 上最优，Athena-PRM 在其余三个子集保持领先。
\textbf{Performance Analysis.} Table~\ref{Table:PRMBench} directly controls for the inference interface across the base model and both training stages. The process-untrained base model obtains only 42.48 overall macro F1, showing that global thinking and single-token decoding alone do not produce a reliable process verifier. With only 5.5K supervised step labels, VRPRM-SFT reaches 66.77, within one point of the strongest prior PRM; the RL stage then lifts VRPRM-RL to 69.18, exceeding EVPV-PRM~\cite{wang2026grounding} by 1.72 points and VisualPRM-8B~\cite{wang2025visualprm}, trained on 400K examples, by 7.00. Gains vary across subsets: VRPRM-RL is best on MathVerse and MathVision, while Athena-PRM keeps the lead on the other three.

% 中文：计算开销分析。表 3 在同一推理实现下联合报告 macro F1、延迟和输出长度。未经过程监督的基础模型既最慢也最不准，输出词元超过 VRPRM-RL 的三倍。VRPRM-RL 每样本约 2.1 秒完成打分，无思考模式进一步降至约 0.55 秒，代价是 3.71 分 macro F1，构成可调节的精度--效率权衡。
\textbf{Computational Overhead Analysis.} Table~\ref{tab:efficiency_analysis} jointly reports macro F1, latency, and generation length under the same inference implementation. The process-untrained base model is both the slowest and the least accurate, generating more than three times the output tokens of VRPRM-RL. VRPRM-RL scores each example in about 2.1 seconds, and the no-thinking mode lowers this to roughly 0.55 seconds at a cost of 3.71 F1 points, exposing a controllable accuracy--efficiency trade-off.

\begin{table}[ht]
    \centering
    \scriptsize
    \setlength\tabcolsep{2pt}
    \resizebox{\columnwidth}{!}{%
    \begin{tabular}{@{}lrrrr@{}}
        \toprule

        \textbf{Model} & \textbf{\shortstack{VPB\\F1}} & \textbf{\shortstack{Time /\\Sample (s)}} & \textbf{\shortstack{Output Tok. /\\Sample}} & \textbf{\shortstack{Output Tok. /\\Step}} \\
        \midrule
        Qwen3-VL-8B-Thinking & 42.48 & 5.466 & 854.34 & 85.97 \\
        \textbf{VRPRM-SFT} & 66.77 & 2.417 & 273.22 & 27.49 \\
        \quad w/o Thinking & 61.63 & 0.999 & 35.09 & 3.53 \\
        \textbf{VRPRM-RL} & \textbf{69.18} & 2.115 & 242.06 & 24.36 \\
        \quad w/o Thinking & 65.47 & 0.551 & 35.00 & 3.52 \\
        \midrule
        VisualPRM-8B & 62.18 & \textbf{0.405} & \textbf{9.94} & \textbf{1.00} \\
        \bottomrule
    \end{tabular}
    }
    % 中文：VisualProcessBench 上的 macro F1 与计算开销。所有模型采用相同的计时与词元统计口径；输出词元包含全局思考和步骤判断。粗体表示每列最优值，时间与词元列越低越好。
    \caption{\textbf{Macro F1 and computational overhead on VisualProcessBench.} All models use the same timing and token-counting protocol. Output tokens include global thinking and step judgments. \textbf{Bold} marks the best value in each column; lower is better for latency and token counts.}
    \label{tab:efficiency_analysis}
\end{table}

\input{Table4BoN.tex}

\subsection{Best-of-N Evaluation Results}

% 中文：在 Best-of-8 任务中，我们使用 VRPRM 作为评价模型。InternVL2.5 策略模型以温度 0.7 生成 8 个带思维链的候选回答，并选择得分最高者作为最终输出。我们在六个多模态推理基准上报告结果，部分专有模型基线来自 OpenCompass 排行榜。
We use VRPRM as the critic for Best-of-8 evaluation. InternVL2.5~\cite{chen2025expandingperformanceboundariesopensource} generates eight CoT responses at temperature 0.7, and the highest-scoring response is returned. We report results on six multimodal reasoning benchmarks---MathVista~\cite{lu2024mathvista}, MathVision~\cite{wang2024measuring}, MathVerse-VO~\cite{zhang2024mathverse}, WeMath~\cite{qiao2024we}, LogicVista~\cite{xiao2024logicvistamultimodalllmlogical}, and MMMU~\cite{yue2024mmmu}; selected proprietary baselines are taken from the OpenCompass leaderboard~\cite{10.1145/3332186.3332253}.

% 中文：表 4 展示了两个训练阶段在三种 InternVL2.5 策略规模上的一致收益。对于 8B、26B 和 38B 策略，VRPRM-SFT 将 Overall 从 36.90、40.62 和 48.88 分别提升至 50.22、54.55 和 62.83，对应 13.32、13.93 和 13.95 分的绝对增益。第二阶段强化学习进一步将 Overall 提升至 58.67、64.04 和 72.00；相对 SFT 再提升 8.45、9.49 和 9.17 分，相对基础策略总计提升 21.77、23.42 和 23.12 分。
Table~\ref{Table:BoN} shows consistent gains from both training stages across all three InternVL2.5 policy scales. For the 8B, 26B, and 38B policies, VRPRM-SFT raises Overall from 36.90, 40.62, and 48.88 to 50.22, 54.55, and 62.83, yielding gains of 13.32, 13.93, and 13.95 points, respectively. VRPRM-RL further raises Overall to 58.67, 64.04, and 72.00---another 8.45, 9.49, and 9.17 points over SFT, and total gains of 21.77, 23.42, and 23.12 points over the base policies.

% 中文：VRPRM-RL 在所有“策略规模--数据集”组合上均优于对应的 VRPRM-SFT，并在 LogicVista 上相对基础策略提升 37.81、35.35 和 33.78 分，在 MathVision 上提升 23.85、23.85 和 27.11 分。相比之下，VisualPRM 的 Overall 仅提升 6.12、6.42 和 3.83 分。InternVL2.5-38B 结合 VRPRM-RL 取得最高 Overall 72.00，较最佳专有模型 Gemini-2.0-Flash 高 16.77 分，说明 RL 学到的过程判断能够稳定迁移到不同策略规模，尤其适用于视觉和逻辑密集任务。
VRPRM-RL improves over its SFT counterpart on all policy-scale--benchmark combinations. Relative to the base policies, its gains on LogicVista are 37.81, 35.35, and 33.78 points, while those on MathVision are 23.85, 23.85, and 27.11 points for the 8B, 26B, and 38B policies, respectively. By comparison, VisualPRM improves Overall by only 6.12, 6.42, and 3.83 points. InternVL2.5-38B with VRPRM-RL achieves the best Overall score of 72.00, exceeding the strongest proprietary model, Gemini-2.0-Flash, by 16.77 points. These results indicate that RL-trained process judgments transfer reliably across policy scales, particularly on visually and logically demanding tasks.

\subsection{Ablation Studies}

\subsubsection{Effects of BoN}

% 中文：为评估测试时扩展能力，我们使用 InternVL2.5-8B 作为策略模型，并在 LogicVista、MathVerse-VO、MathVista 和 MathVision 上比较 VRPRM-SFT、VRPRM-RL、VisualPRM 与多数投票。
To evaluate test-time scaling, we use InternVL2.5-8B as the policy and compare VRPRM-SFT, VRPRM-RL, VisualPRM, and majority voting on LogicVista, MathVerse-VO, MathVista, and MathVision.

\begin{figure*}[t!]
    \centering
    \includegraphics[width=0.77\textwidth]{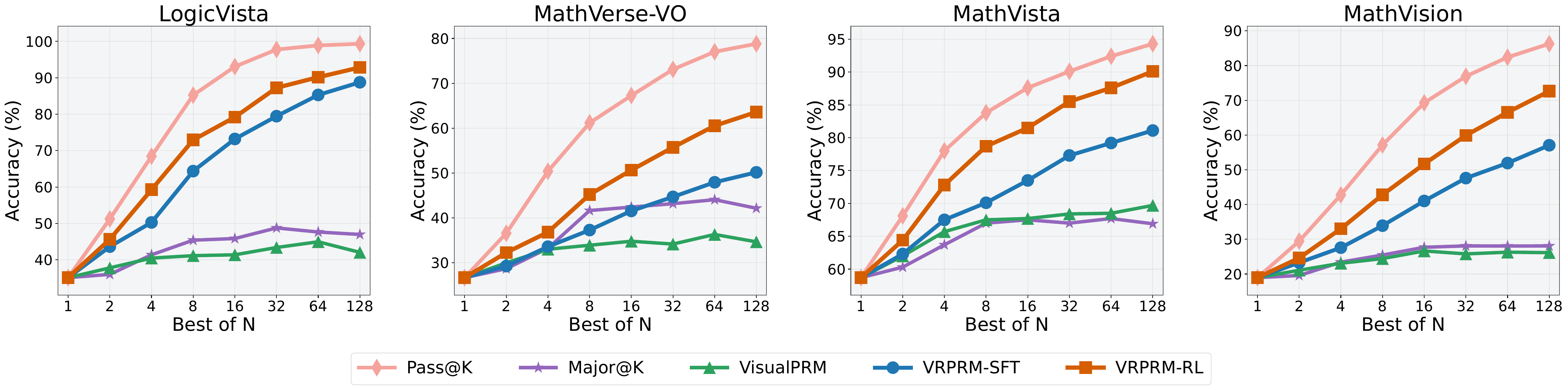}
    % 中文：InternVL2.5-8B 在四个多模态推理基准上的 Best-of-N 结果，对比 VRPRM-SFT、VRPRM-RL、VisualPRM 和多数投票；Pass@K 为理论上界。
    \caption{Best-of-\(N\) results for InternVL2.5-8B on four benchmarks. Pass@K is the oracle upper bound; Major@K is majority voting.}
    \label{fig:bon}
\end{figure*}

% 中文：图 3 展示测试时扩展趋势。VisualPRM 与多数投票随着候选数增加很快趋于饱和，而两个 VRPRM 检查点均能持续利用更多候选。除所有方法相同的 Bo1 起点外，VRPRM-RL 在四项基准和全部 \(N\) 上均优于 VRPRM-SFT；到 Bo128 时，RL 相对 SFT 在 MathVista、MathVision、MathVerse-VO 和 LogicVista 上分别领先 9.00、15.56、13.44 和 4.08 分。这说明第二阶段 RL 不仅提高固定 Bo8 的选择准确率，也增强了随候选规模增长的排序能力。
Figure~\ref{fig:bon} shows the test-time scaling trends. VisualPRM and majority voting quickly saturate as the candidate pool grows, whereas both VRPRM checkpoints continue to exploit additional candidates. Beyond their shared Bo1 starting point, VRPRM-RL outperforms VRPRM-SFT for every \(N\) on all four benchmarks. At Bo128, RL leads SFT by 9.00, 15.56, 13.44, and 4.08 points on MathVista, MathVision, MathVerse-VO, and LogicVista, respectively. Thus, second-stage RL improves not only fixed-Bo8 selection but also ranking quality as the candidate pool scales.

\subsubsection{Effects of Global Thinking}

% 中文：我们关闭全局 thinking 阶段：单次请求包含问题、完整候选解答和图像，一次性返回全部步骤得分的 JSON 数组。两个变体因此每个样本恰好看到一次相同的视觉输入，从而隔离显式推理阶段对准确率和计算开销的贡献。
We disable the global-thinking stage: one request with the question, full candidate solution, and images returns all step scores as a JSON array. Both variants thus see the same visual input exactly once per example, isolating the explicit reasoning stage's contribution to accuracy and computation.

% 中文：性能收益。关闭 thinking 后，VRPRM-SFT 的总体 macro F1 从 66.77 降至 61.63，下降 5.14 分；VRPRM-RL 从 69.18 降至 65.47，下降 3.71 分。该结果表明，全局视觉推理在两个训练阶段都能稳定改善步骤判别，与"其有助于理解步骤间依赖关系"的假设一致；同时，无 thinking 的模型仍明显优于 42.48 的基础模型，说明过程监督本身与显式推理均是有效增益来源。
\textbf{Performance Gain.} Disabling thinking reduces overall macro F1 from 66.77 to 61.63 for VRPRM-SFT, a 5.14-point drop, and from 69.18 to 65.47 for VRPRM-RL, a 3.71-point drop. Global visual reasoning therefore consistently improves step discrimination at both stages, consistent with the hypothesis that it helps resolve dependencies among candidate steps. Meanwhile, both non-thinking variants remain well above the 42.48 base model, showing that process supervision and explicit reasoning provide complementary gains.

% 中文：计算开销。对 VRPRM-SFT，关闭 thinking 将平均输出长度从 273.22 降至 35.09 个词元，并将单样本时间从 2.417 秒降至 0.999 秒；对 VRPRM-RL，相应数值从 242.06 降至 35.00 个词元、从 2.115 秒降至 0.551 秒。因此，可根据应用场景在完整推理模式与低延迟模式之间选择。完整推理模式还能提供可检查的全局判断依据；补充材料中的评分案例展示了模型如何归纳视觉规则、定位首个错误并拒绝依赖该错误的后续结论。
\textbf{Computational Overhead.} For VRPRM-SFT, disabling thinking reduces mean output length from 273.22 to 35.09 tokens and latency from 2.417 to 0.999 seconds per example. For VRPRM-RL, the corresponding values decrease from 242.06 to 35.00 tokens and from 2.115 to 0.551 seconds. The model can thus trade a modest amount of accuracy for a substantially lower-cost deployment mode. The full reasoning mode additionally exposes an inspectable global rationale; the scoring example in Appendix~\ref{app:scoring_example} shows how VRPRM induces a visual rule, localizes the first error, and rejects conclusions that depend on it.

% 中文：总体而言，全局思考虽增加计算开销，却显著提升奖励建模效果；这可以合理解释为其帮助模型理解步骤间的因果与逻辑关系，但该机制解释仍属假设，有待针对性实验验证。
Overall, global thinking introduces a clear computational trade-off but substantially improves reward-model accuracy, plausibly by helping the critic resolve causal and logical relationships between steps.

\subsubsection{Effects of RL}

% 中文：我们通过 continued-SFT 对照隔离第二阶段优化目标的作用：从同一 VRPRM-SFT 检查点出发、在同一 40K 全高置信池上继续监督训练，并与 VRPRM-RL 在 VisualProcessBench 步骤判断（表 2）和 Best-of-8 选择（表 4）上比较。
We isolate the second-stage objective with a continued-SFT control trained from the same VRPRM-SFT checkpoint on the same 40K high-confidence pool, compared with VRPRM-RL on VisualProcessBench step judgments (Table~\ref{Table:PRMBench}) and Best-of-8 selection (Table~\ref{Table:BoN}).

% 中文：在同一数据池上继续 SFT 使 VisualProcessBench Overall 从 66.77 降至 61.22，每个子集都低于 SFT 起点；而 RL 达到 69.18（相对 SFT 提升 2.41 分）。两个第二阶段消耗相同的标签池，因此 7.96 分的差距将增益归因于 RL 目标本身，而非额外的监督数据或优化。关闭 thinking 时，RL 仍将 Overall 从 61.63 提高到 65.47，说明其直接改善步骤判别与候选排序，而非仅优化思维文本。
\looseness=1
Continued SFT on the same pool degrades Overall from 66.77 to 61.22, below the SFT starting point on every subset, whereas RL reaches 69.18. Both second stages consume the same label pool, so the 7.96-point gap attributes the gain to the RL objective rather than additional supervision or optimization. Without thinking, RL still raises Overall from 61.63 to 65.47, showing that it directly improves step discrimination and candidate ranking rather than merely optimizing rationale generation.

% 中文：Best-of-8 Overall 在 8B、26B 和 38B 策略上分别提升 8.45、9.49 和 9.17 分，覆盖全部 18 个“策略规模--数据集”组合。因此，在高置信的非 CoT 数据上进行 RL 无需额外 CoT 标注即可增强过程判断，同时保留 SFT 建立的全局推理范式。
Best-of-8 Overall improves by 8.45, 9.49, and 9.17 points for the three policies, with gains on all 18 policy-scale--benchmark combinations. Thus, RL on high-confidence non-CoT data strengthens process judgments without additional CoT annotation while preserving the global-reasoning behavior established by SFT.

\section{Conclusion}

% 中文：本文提出 VRPRM，将少量质量受控、评测对齐的 CoT-PRM 监督与高置信的非 CoT 过程级强化学习相结合。仅使用 5.5K 个可训练步骤标签和 40K 条 RL 样本，VRPRM 在 VisualProcessBench 上达到 69.18 macro F1，较此前最强 PRM 高 1.72 分，并比使用 400K 样本训练的 VisualPRM-8B 高 7.00 分。作为 Best-of-8 评价模型，VRPRM-RL 在三种策略规模上带来 21.77--23.42 分的绝对提升，表明该方法具有良好的标注效率和测试时扩展能力。两个训练阶段具有互补作用：加权 SFT 建立可复用的全局视觉分析和严格对齐的步骤判断，RL 则无需额外 CoT 标注即可增强类别平衡的步骤判别与输出完整性。该方法在所有 VisualProcessBench 子集以及全部 18 个“策略规模--基准”组合上均取得一致提升，说明其收益并非局限于单一任务或策略规模。除准确率外，先生成一次全局分析、再逐步输出严格对齐的二值判断，能够避免步骤数量不匹配并提供可检查的评价依据。目前的监督主要来自 MCTS 分数且集中于视觉数学与逻辑推理；未来工作将研究该训练方案在更开放的视觉任务、人工验证步骤标签和安全关键过程评价中的迁移能力。
We introduced VRPRM, combining a small amount of quality-controlled, evaluation-aligned CoT-PRM supervision with process-level RL on high-confidence non-CoT annotations. Using 5.5K trainable step labels and 40K RL examples, VRPRM reaches 69.18 macro F1 on VisualProcessBench, exceeding the strongest prior PRM by 1.72 points and VisualPRM-8B trained on 400K examples by 7.00 points. As a Best-of-8 critic, VRPRM-RL yields absolute gains of 21.77--23.42 points across three policy scales, demonstrating strong label efficiency and test-time scaling. The two stages are complementary: weighted SFT establishes reusable global analysis and aligned step judgments, whereas RL improves class-balanced discrimination and output completeness without further CoT annotation.

\section*{Limitations}
% 中文：本节讨论本文的两个主要局限。其一，全部训练监督来自 MCTS 估计的步骤分数而非人工验证的判断：双阈值过滤与教师一致性校验可缓解歧义，但 MC 估计的噪声与人工选定的阈值（0.125/0.75，两阶段共用）仍可能传入监督信号；最终步骤标签未做人工核验，人工审计留作未来工作。相应地，效率声明限定为标注效率（以少得多的标注取得更高精度），与 VisualPRM-8B（400K 样本、不同骨干）的对比并非受控的数据预算对照。其二，显式全局思考以推理开销换取验证精度：带思考打分每样本需 2.115 秒与 242.06 个输出词元（无思考为 0.551 秒与 35.00 个），部署方需按精度--效率需求在完整推理与低延迟模式之间取舍。
This work has two main limitations. First, all training supervision derives from Monte Carlo-estimated step scores in VisualPRM400K-v1.1-Raw rather than human-verified judgments. Dual-threshold filtering and teacher-consistency checks mitigate ambiguity, but noise in the Monte Carlo estimates and the hand-chosen thresholds (0.125/0.75, shared by both training stages) can still propagate into the supervision signal; the final step labels are not human-verified, and a manual audit is left as future work. Accordingly, our efficiency claim is limited to label efficiency---higher accuracy from far fewer annotations---since the comparison with VisualPRM-8B (400K examples, a different backbone) is not a controlled data-budget contrast. Second, explicit global thinking trades verification accuracy for inference cost: scoring with thinking takes 2.115 seconds and 242.06 output tokens per example, versus 0.551 seconds and 35.00 tokens without thinking, so deployments must choose between the full-reasoning and low-latency modes according to their accuracy--efficiency requirements.

\bibliography{vrprm}

\appendix

\section{Kimi-K2.6 Rollout and Multi-Turn SFT Construction}
\label{Appendix:RollPrompt}
% 中文：本节详细说明教师模型 rollout、置信标签筛选、多轮序列化和最终 SFT 数据统计。
This section details teacher rollout, confidence-based label filtering, multi-turn serialization, and the resulting SFT data statistics.

\subsection{Teacher Rollout and Quality Control}
% 中文：对于源数据中的第 \(i\) 个蒙特卡洛得分 \(q_i\)，我们设置负阈值 \(\tau_-=0.125\) 和正阈值 \(\tau_+=0.75\)。当 \(q_i\leq\tau_-\) 时标签为 0，当 \(q_i\geq\tau_+\) 时标签为 1，其余步骤记为不确定。至少包含一个负步骤的样本进入负源池；所有步骤均为正的样本进入正源池；正源池提供完整正确过程，避免模型过度预测 0。
For a source Monte Carlo score \(q_i\), we use a negative threshold \(\tau_-=0.125\) and a positive threshold \(\tau_+=0.75\). We assign label 0 when \(q_i\leq\tau_-\), label 1 when \(q_i\geq\tau_+\), and mark all intermediate values as uncertain. We form a negative-source pool from examples containing at least one negative step and a positive-source pool from examples whose steps are all positive; examples satisfying neither condition are not selected. The positive pool supplies complete correct trajectories and prevents the model from becoming biased toward predicting negative labels.

% 中文：Kimi-K2.6 的首次调用同时接收图像、问题、参考答案和完整候选解答。模型在隐藏推理通道中分析关键视觉证据、步骤依赖和首个关键错误，并在可见通道中仅返回严格 JSON 数组。数组长度必须与候选步骤数完全一致，且所有高置信位置必须与蒙特卡洛标签一致，否则样本被拒绝并重试。
The first Kimi-K2.6 call receives the image, problem, reference answer, and complete candidate solution. In its hidden reasoning channel, the teacher analyzes key visual evidence, step dependencies, and the first consequential error; its visible response contains only a strict JSON score array. We require the array length to exactly match the number of candidate steps and every confident position to agree with the Monte Carlo label; otherwise, the sample is rejected and retried.

% 中文：随后使用第二次确定性调用将教师的隐藏推理压缩为最多 180 个词、100 至 2000 个字符的可见全局思考。压缩结果必须只包含一个有效的 <think>...</think> 块，过短、缺失或含占位符的结果都会被过滤。
A second deterministic call compresses the teacher's hidden reasoning into visible global thinking of at most 180 words and between 100 and 2,000 characters. The result must form one valid \texttt{<think>... </think>} block; missing, overly short, or placeholder outputs are filtered.

% 中文：教师模型通过 OpenAI 兼容 API 端点访问（Kimi-K2.6），两次调用的完整 system prompt 见图 \ref{fig:teacher-prompts}。第一次调用温度 0.2、最多 2,048 输出词元，保持原生 thinking 开启，其隐藏推理保留在 API 的 reasoning 字段中供第二次调用复用；第二次调用为确定性设置（温度 0、最多 512 词元、经 chat template 关闭 thinking），输入为完整问题文本与截断至 12,000 字符的教师隐藏推理，不含图像。两次调用作为整体每样本最多完整重试 3 次（线性退避 2/4 秒）。双源池合计接受率 65.7%（负池 46.8%、正池 82.8%），重试耗尽的样本记为失败并剔除。
The teacher is accessed as Kimi-K2.6 behind an OpenAI-compatible API endpoint; the verbatim system prompts of both calls are given in Figure~\ref{fig:teacher-prompts}. The first call uses temperature 0.2 and at most 2,048 output tokens with the native thinking mode left on; its hidden reasoning is preserved in the API reasoning field and reused by the second call. The second call is deterministic---temperature 0, at most 512 tokens, thinking disabled through the chat template---and receives the full problem text together with the teacher's hidden reasoning truncated to 12,000 characters, without images. The two calls form one atomic unit retried at most three times per sample with linear backoff (2 and 4 seconds). Overall acceptance is 65.7\% across the two source pools (46.8\% for the negative pool, 82.8\% for the positive pool); samples that exhaust the retry budget are recorded as failures and excluded.

\begin{figure*}[t]
\begin{tcolorbox}[title=Verbatim Kimi-K2.6 Teacher Prompts, width=\textwidth, fonttitle=\bfseries]
\small
\textbf{Call 1 --- system prompt} (hidden-reasoning analysis and step scoring; temperature 0.2, max 2,048 tokens, thinking on):\\[2pt]
You are a visual process reward model teacher. Analyze the image(s), question, reference answer, and complete candidate solution before any step-level supervision is created.\\
Focus your internal reasoning on:
\begin{itemize}\setlength\itemsep{0pt}\setlength\parskip{0pt}
\item key visual evidence needed to solve the problem;
\item the question goal and reference-answer constraint;
\item dependencies between candidate steps;
\item the first consequential incorrect step if one exists;
\item later steps that are correct independently or are invalid because they rely on an earlier error.
\end{itemize}
After reasoning, output strict JSON only:\\
\texttt{\{"Score": [one integer 0 or 1 for each candidate step]\}}\\
The Score array length must exactly equal the number of candidate steps.\\[4pt]
\textbf{Call 1 --- user input:} [Question], [Reference Answer], and [Candidate Solution] (Step 0 to Step $n$, separated by newlines), with all images attached.\\[4pt]
\textbf{Call 2 --- system prompt} (deterministic compression into global thinking; temperature 0, max 512 tokens, thinking disabled):\\[2pt]
You compress a teacher model's hidden reasoning into one concise global thinking block for SFT.\\
Output requirements:
\begin{itemize}\setlength\itemsep{0pt}\setlength\parskip{0pt}
\item Output exactly one \texttt{<think></think>} block.
\item Preserve the key visual evidence, question goal, reference-answer constraint, dependencies between steps, and the main correctness issue.
\item Do not output step scores.
\item Do not output JSON.
\item Keep the thinking focused and no more than 180 words.
\end{itemize}
\textbf{Call 2 --- user input:} the full problem text followed by [Teacher Hidden Reasoning] (truncated to 12,000 characters), without images.
\end{tcolorbox}
% 中文：Kimi-K2.6 两次调用的逐字 system prompt 与用户输入构成。
\caption{Verbatim system prompts and user-input composition for the two Kimi-K2.6 teacher calls.}
\label{fig:teacher-prompts}
\end{figure*}

\subsection{Evaluation-Aligned Multi-Turn Serialization}
% 中文：通过筛选后，同一源样本的全局思考和所有步骤判断被合并为一条对话。首个用户轮次包含完整问题上下文，助手生成全局思考；之后每个用户轮次仅指出当前步骤和此前判断，助手只返回 0 或 1。该结构与评测时的轮次组织和输出约束保持一致，并保证步骤数量由请求轮次数确定，而不是依赖模型一次性生成数组。
After filtering, the global thinking and all step judgments from one source example are merged into a single conversation. The first user turn contains the complete problem context and elicits global thinking. Each later user turn specifies the current step and previous judgments, and the assistant returns only 0 or 1. This structure matches the evaluation-time turn organization and output constraints, and determines the number of judgments from the requested turns rather than from a model-generated array.

\begin{figure*}[t]
    \centering
    \begin{tcolorbox}[title=Evaluation-Aligned Multi-Turn SFT Format, width=\textwidth,
    fonttitle=\bfseries]
    \textbf{System:} Generate global visual reasoning once; answer subsequent scoring turns with exactly one token, 0 or 1.\\
    \textbf{User 0:} [Image(s)] [Question] [Reference Answer] [Complete Candidate Solution]\\
    \textbf{Assistant 0:} \texttt{<think>} concise global visual reasoning \texttt{</think>}
    \hfill \textit{loss scale = 0.5}\\
    \textbf{User \(i\):} [Previous Step Judgments] [Current Step \(s_i\)]\\
    \textbf{Assistant \(i\):} \texttt{0} or \texttt{1}
    \hfill \textit{loss scale = 2.0 if confident; loss = false otherwise}
    \end{tcolorbox}
    % 中文：与评测方式对齐的多轮 SFT 样本结构。全局思考只生成一次；每个后续助手轮次仅预测一个步骤标签。
    \caption{Evaluation-aligned multi-turn SFT format. Global thinking is generated once, and each subsequent assistant turn predicts exactly one step label.}
    \label{fig:multiturn-sft}
\end{figure*}

% 中文：中间置信区间中的步骤不会被删除。我们使用 Kimi 的二值判断填充该助手轮次，使后续步骤仍可读取完整历史，但设置 loss=false 将其排除出监督目标。对于高置信步骤，训练目标始终使用源标签；同时按步骤而非样本进行采样，使最终可训练标签中的负例比例接近 35%。
Uncertain steps are not removed. We use the Kimi binary judgment for that assistant turn so later steps retain a complete history, but set \texttt{loss=false} to exclude it from supervision. Confident turns always use the source label. We additionally balance at the step level, targeting a 35\% negative ratio among trainable judgments.

\subsection{Dataset Statistics}
% 中文：我们直接统计最终发布的多轮 SFT JSON。所有 1378 条对话均使用相同阈值：当源分数 \(q_i\leq0.125\) 时步骤标为负例，当 \(q_i\geq0.75\) 时标为正例，其余步骤保留为历史上下文但屏蔽损失。该阈值设计在保留完整推理历史的同时，避免决策边界附近的模糊分数进入监督目标。
We compute all statistics directly from the final multi-turn SFT JSON. Every one of the 1,378 conversations uses the same thresholds: a source score \(q_i\leq0.125\) is labeled negative, \(q_i\geq0.75\) is labeled positive, and intermediate scores remain in the dialogue history with their loss masked. This preserves complete reasoning context while preventing ambiguous boundary scores from entering the supervised objective.

% 中文：最终数据包含 1,378 条源样本级多轮对话（1,364 条用于训练、14 条用于验证），来自 38 个源标注文件及 1,332 张不同图像。每条对话由系统消息、首轮问题上下文用户消息、一次全局思考助手消息以及每个候选步骤一对用户-助手消息构成，共计 18,616 条消息；这些消息覆盖 7,241 个候选步骤，其中 5,592 个高置信步骤（77.2\%）作为监督标签参与训练，1,649 个不确定步骤（22.8\%）仅作为屏蔽损失的上下文保留。可训练标签中包含 1,954 个负例和 3,638 个正例，负例比例为 34.9\%。全局思考长度的中位数为 779 个字符。
The resulting dataset contains 1,378 source-level multi-turn conversations---1,364 used for training and 14 held out for validation---drawn from 38 source annotation files and 1,332 distinct images. Each conversation consists of a system message, an initial user turn providing the problem context, one global-thinking assistant turn, and a user--assistant pair for every candidate step, for a total of 18,616 messages. These cover 7,241 candidate steps, of which 5,592 confident steps (77.2\%) serve as supervised labels and 1,649 uncertain steps (22.8\%) are retained only as loss-masked context. The trainable labels comprise 1,954 negatives and 3,638 positives, yielding a 34.9\% negative ratio. The median global-thinking length is 779 characters.

\begin{table}[h]
    \centering
    \footnotesize
    \setlength{\tabcolsep}{3pt}
    \begin{tabular}{lr}
        \toprule
        Statistic & Count \\
        \midrule
        Source-level conversations & 1,378 \\
        \quad train / validation conversations & 1,364 / 14 \\
        Source annotation files / unique images & 38 / 1,332 \\
        Negative / positive source samples & 910 / 468 \\
        Confidence thresholds \(\tau_- / \tau_+\) & 0.125 / 0.75 \\
        Global-thinking turns & 1,378 \\
        Median global-thinking length (characters) & 779 \\
        Total messages & 18,616 \\
        Candidate steps & 7,241 \\
        Trainable (supervised) steps & 5,592 \\
        Masked uncertain steps & 1,649 \\
        Negative / positive trainable labels & 1,954 / 3,638 \\
        \bottomrule
    \end{tabular}
    % 中文：最终多轮 SFT 数据集统计。正负比例按可训练步骤标签计算，而非按源样本计算。
    \caption{Statistics of the final multi-turn SFT dataset. Class balance is computed over trainable step labels rather than source examples.}
    \label{tab:multiturn-sft-stats}
\end{table}

\begin{figure*}[t]
    \centering
    \includegraphics[width=\textwidth]{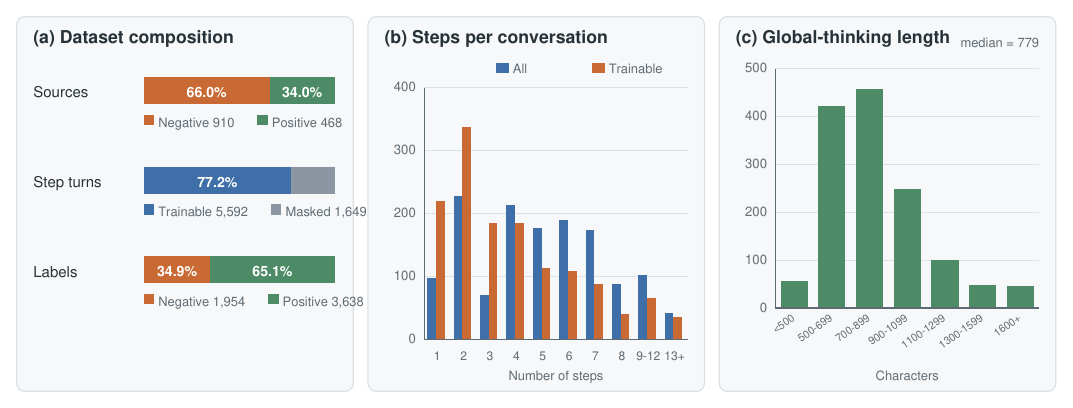}
    % 中文：最终多轮 SFT 数据集的统计分布。（a）源样本极性、步骤监督状态和可训练标签类别构成；（b）每条对话中全部步骤数与可训练步骤数的分布；（c）全局思考字符长度分布。
    \caption{\textbf{Statistics of the final multi-turn SFT dataset.} (a) Composition by source polarity, step supervision status, and trainable label class. (b) Distributions of all and trainable steps per conversation. (c) Character-length distribution of the global-thinking targets.}
    \label{fig:sft-dataset-statistics}
\end{figure*}

\section{Confidence-Filtered RL Data and Reward Specification}
\label{Appendix:RLReward}

\subsection{RL Data Construction}
% 中文：RL 数据从 VisualPRM400K-v1.1-Raw 构建。每条原始样本提供问题、图像、参考答案、完整候选步骤以及 MCTS 得到的连续步骤奖励。对于第 \(i\) 个源分数 \(q_i\)，当 \(q_i\leq0.125\) 时标记为 0，当 \(q_i\geq0.75\) 时标记为 1；若任一步骤满足 \(0.125<q_i<0.75\)，则整条源样本被过滤。该样本级过滤保留了完整步骤索引，并确保奖励不受模糊中间分数污染。
We construct the RL data from VisualPRM400K-v1.1-Raw. Each raw example provides a question, image, reference answer, complete candidate steps, and continuous MCTS-derived step rewards. We assign label 0 when \(q_i\leq0.125\) and label 1 when \(q_i\geq0.75\). If any step satisfies \(0.125<q_i<0.75\), the entire source example is filtered. This example-level rule preserves complete step indexing and prevents ambiguous intermediate scores from contaminating the reward. Throughout the paper, ``high-confidence'' refers to this filtering rule over Monte Carlo estimates rather than human-verified correctness.

% 中文：最终数据集包含 40,000 条训练样本和 2,000 条留出验证样本。每条样本包含 prompt、images 和 ground_truth 三个顶层字段；ground_truth 以结构化形式保存源标识、原始答案与标注、源步骤分数、二值步骤标签、正负步骤索引、类别以及实际使用的正负阈值，从而支持严格对齐、类别分析与数据溯源。
The final dataset contains 40,000 training examples and 2,000 held-out validation examples. Each has three top-level fields: \texttt{prompt}, \texttt{images}, and \texttt{ground\_truth}. The structured ground truth stores source identifiers, the original answer and annotation, source step scores, binary step labels, positive and negative indices, category, and the explicit positive and negative thresholds, supporting exact alignment, category analysis, and provenance.

\begin{table}[t]
    \centering
    \small
    \begin{tabular}{lr}
        \toprule
        \textbf{RL dataset statistic} & \textbf{Value} \\
        \midrule
        Training / validation examples & 40,000 / 2,000 \\
        Negative / positive source examples & 22,658 / 17,342 \\
        Negative / positive training steps & 52,774 / 112,227 \\
        Total training step labels & 165,001 \\
        Negative-step ratio & 31.98\% \\
        Steps per example & 1--24 \\
        \bottomrule
    \end{tabular}
    % 中文：置信度过滤并平衡采样后的 RL 数据统计。源样本极性与步骤极性分别统计。
    \caption{Statistics of the confidence-filtered and balanced RL data. Source polarity and step-label polarity are reported separately.}
    \label{tab:rl-data-stats}
\end{table}

\begin{figure*}[t]
\centering
\begin{tcolorbox}[title=RL Prompt and Constrained Rollout Format, fonttitle=\bfseries]
\begin{minipage}[t]{0.43\textwidth}
\small
\textbf{Training prompt}\\[2pt]
\texttt{[Question]}\\
\texttt{...}\\[2pt]
\texttt{[Reference Answer]}\\
\texttt{...}\\[2pt]
\texttt{[Candidate Solution Steps]}\\
\texttt{Step 0: ...}\\
\texttt{Step 1: ...}\\[2pt]
\texttt{[Task]}\\
First write one concise \texttt{<think>...</think>} block.\\
Then write one \texttt{<answer>...</answer>} block.\\[8pt]
\textbf{Required rollout}\\[2pt]
\texttt{<think>}\\
Compact global visual reasoning.\\
\texttt{</think>}\\
\texttt{<answer>}\\
\texttt{Step 0: 1}\\
\texttt{Step 1: 0}\\
\texttt{...}\\
\texttt{</answer>}
\end{minipage}
\hfill
\begin{minipage}[t]{0.54\textwidth}
\small
\textbf{System prompt}\\[2pt]
\texttt{\{\{ content | trim \}\}}\\[2pt]
You are a visual process reward model. First generate global thinking, then score every candidate solution step.\\[3pt]
Output exactly in this format: one \texttt{<think>...</think>} block containing compact global reasoning about the images, question goal, reference-answer constraint, step dependencies, and the likely first consequential error; then one \texttt{<answer>...</answer>} block containing only indexed binary labels.\\[3pt]
\textbf{Scoring policy:}
\begin{itemize}
    \setlength{\itemsep}{0pt}
    \setlength{\topsep}{2pt}
    \item Output one score for every candidate step using the prompt indices.
    \item Put final scores only in \texttt{<answer>} as \texttt{Step i: 0/1}.
    \item Use 1 only when the step is fully supported and remains logically and mathematically correct.
    \item Use 0 for unsupported, visually mistaken, logically invalid, computationally wrong, contradictory, or unrecovered dependent steps.
    \item Do not output text outside the two required blocks.
\end{itemize}
\end{minipage}
\end{tcolorbox}
% 中文：RL 用户 prompt、system prompt 和受约束 rollout 的结构。输出只允许一个全局思考块和一个包含按索引排列二元步骤标签的答案块。
\caption{RL user prompt, system instruction, and constrained rollout structure. The output contains exactly one global-thinking block and one answer block with indexed binary step labels.}
\label{fig:rl-prompt-format}
\end{figure*}

\subsection{Reward Parsing and Decomposition}
% 中文：奖励解析器首先尝试严格匹配唯一且非空的 <think> 与 <answer> 块，且不允许块外文本；满足时格式得分为 1，严格结构中存在空块时为 0.5。对于非严格输出，解析器分别对存在非空 think（0.30）、存在非空 answer（0.40）、输出以 <think> 开始（0.15）和包含 </answer> 结尾标记（0.15）给部分分，最高截断为 0.8。该分档设计为格式漂移保留连续学习信号。
The parser first attempts a strict match of one nonempty \texttt{<think>} block followed by one nonempty \texttt{<answer>} block, with no outside text; this receives \(\mathcal{R}_{\mathrm{format}}=1\), while a strict structure with an empty block receives 0.5. For non-strict outputs, it adds partial credit for nonempty thinking (0.30), a nonempty answer (0.40), beginning with \texttt{<think>} (0.15), and containing the closing \texttt{</answer>} tag (0.15), capped at 0.8. This graded design preserves a learning signal under format drift.

% 中文：尽管 prompt 强制使用 “Step i: 0/1”，步骤解析器还能够容忍带项目符号的等价写法，以及二元列表或字典，以避免可恢复的表面错误完全丢失训练信号。结构得分与任务正确性相互独立；无法解析出步骤标签时，步骤奖励和数量一致性仍会显著下降，因此仅有形式正确的 answer 无法获得较高总奖励。
Although the prompt enforces ``Step \(i\): 0/1,'' the step parser also accepts equivalent bulleted lines and binary JSON lists or dictionaries, avoiding the loss of recoverable supervision. Structural compliance is separated from task correctness: if no step labels are parseable, both the step reward and count consistency collapse, so a merely well-formed answer cannot receive a high total reward.

% 中文：步骤得分在每个源样本内部计算。令 \(\mathcal{C}\subseteq\{0,1\}\) 为真实标签或多余预测中出现的类别集合，则 source macro F1 对这些有效类别的 F1 取平均。该定义在单一类别样本上不会人为引入一个不存在的类别。
The step score is computed independently within each source example. Let \(\mathcal{C}\subseteq\{0,1\}\) contain classes represented by target labels or extra predictions. The source-level macro F1 averages over these active classes, avoiding an artificial absent class in single-class examples:
\begin{equation}
\operatorname{F1}_{\mathrm{macro}}
=\frac{1}{|\mathcal{C}|}\sum_{c\in\mathcal{C}}\operatorname{F1}_{c}.
\end{equation}
% 中文：该目标同时惩罚漏检错误步骤和将正确步骤误判为错误，避免类别不均衡时被单一类别准确率主导。令真实索引集合为 \(I=\{0,\ldots,n-1\}\)，解析出的索引集合为 \(\widehat I\)，则多余与缺失索引共同决定数量得分。实际训练配置先用该数量得分截断原始 macro F1，再以 0.85 和 0.15 的权重组合两项。
This objective penalizes both missed errors and false rejection of valid steps, preventing a dominant class from controlling the reward. For target indices \(I=\{0,\ldots,n-1\}\) and parsed indices \(\widehat I\), extra and missing indices jointly determine
\begin{equation}
\begin{aligned}
\mathcal{R}_{\mathrm{count}}
&=\max\!\left(0,1-\tfrac{|\widehat I\setminus I|+|I\setminus\widehat I|}{\max(n,1)}\right),\\
\mathcal{R}_{\mathrm{step}}
={}&0.85\min\!\left(\operatorname{F1}_{\mathrm{macro}},\mathcal{R}_{\mathrm{count}}\right)\\
&\quad+0.15\,\mathcal{R}_{\mathrm{count}}.
\end{aligned}
\end{equation}

% 中文：该样本内聚合与 VisualProcessBench 评测所用的池化 macro F1（先跨子集池化全部步骤再计算两类 F1，见附录 C.3）对齐但不相同：RL 优化的是逐样本形式的代理目标。
This within-example aggregation is aligned with, but not identical to, the pooled macro F1 used for VisualProcessBench evaluation, which pools all steps across subsets before computing the two-class F1; RL therefore optimizes a per-example surrogate of the evaluation metric.

% 中文：思考长度奖励使用字符数 \(L\)。缺失 think 块时得分为 0；少于 80 个字符时线性增加；80 至 1200 个字符之间保持满分；超过 1200 个字符后线性下降并截断到 0。它奖励的是信息充分且适中的全局思考，而不是越长越好。
The thinking-length reward uses character count \(L\). A missing thinking block receives zero; short reasoning is rewarded linearly up to 80 characters, 80--1,200 characters receive full credit, and longer outputs decay linearly with a floor at zero. Thus, it rewards sufficiently informative but moderate global thinking rather than length itself:
{\small
\[
\mathcal{R}_{\mathrm{think}}(L)=
\begin{cases}
0, & \text{no think block},\\
L/80, & 0\leq L<80,\\
1, & 80\leq L\leq 1200,\\
\multicolumn{2}{@{}l@{}}{\max(0,\,2-L/1200)\;\;(L>1200)}.
\end{cases}
\]
}

% 中文：本次训练通过配置文件覆盖奖励函数的默认参数，实际采用步骤、格式和思考权重 0.92、0.05 和 0.03；三者之和为 1。该比例使每条源样本内的正负类判别始终主导优化，同时利用小权重约束输出结构与思考长度。
The training configuration overrides the reward function defaults and uses weights of 0.92, 0.05, and 0.03 for step correctness, format, and thinking length, respectively. They sum to one, keeping within-source positive/negative discrimination dominant while using small auxiliary terms to stabilize structure and rationale length.

\section{Implementation Details}
\label{Appendix:ImplementationDetails}

\begin{figure*}[!t]
    \centering
    \includegraphics[width=0.97\textwidth]{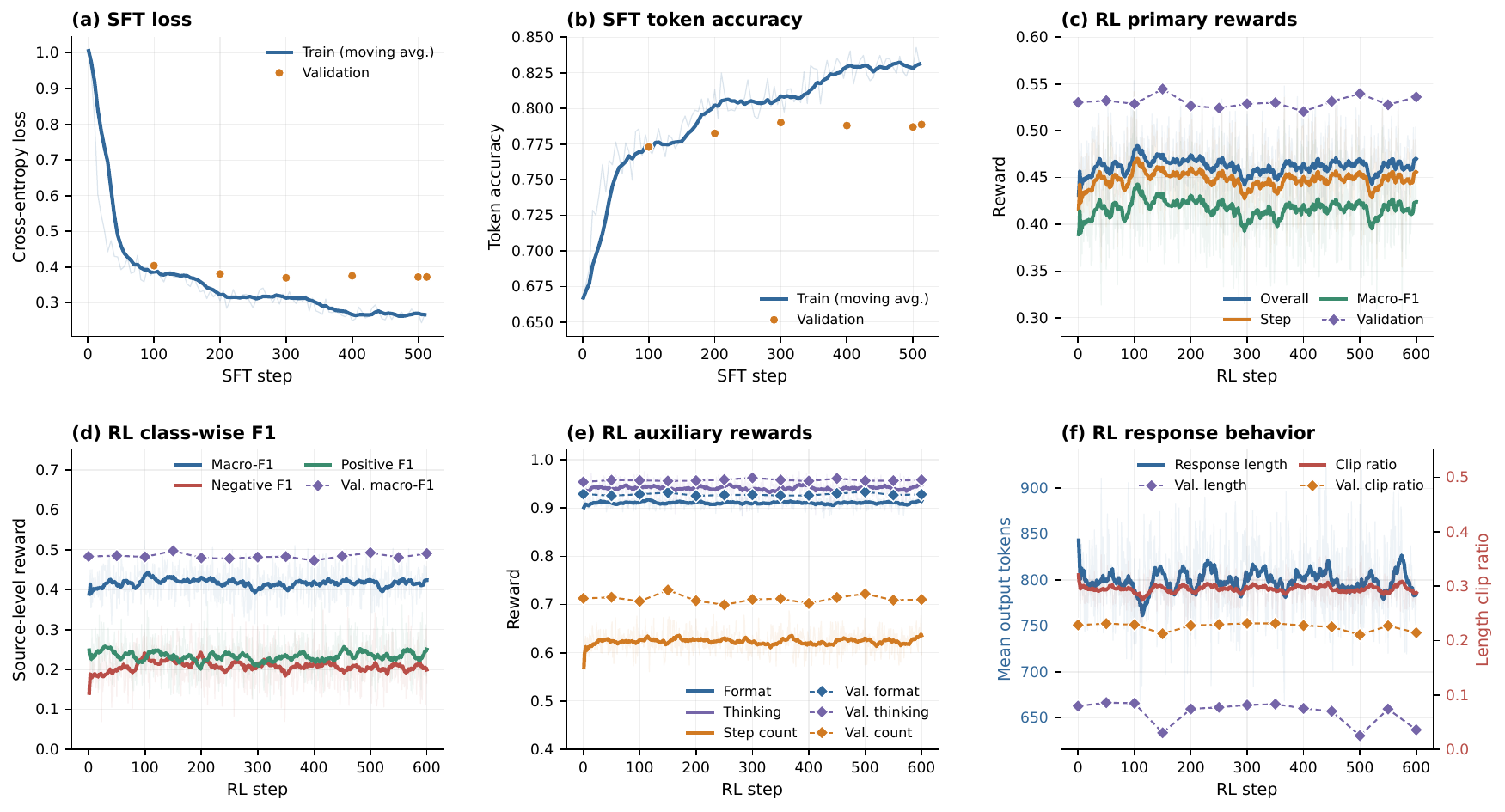}
    % 中文：VRPRM 的两阶段训练曲线。（a--b）SFT 损失与 token accuracy；（c--f）完整 600 步 RL 训练中的主奖励、分类 F1、辅助奖励和回答行为。浅色线表示原始批次值，粗线分别表示 SFT 的因果七步移动平均和 RL 的因果二十一步移动平均；圆点表示 SFT 验证结果，菱形虚线表示每 50 步一次的 RL 验证结果。
    \caption{Two-stage VRPRM training dynamics. (a--b) SFT loss and token accuracy. (c--f) Primary rewards, class-wise F1, auxiliary rewards, and response behavior over the complete 600-step RL run. Faint traces are raw batch values; bold traces are causal seven-step moving averages for SFT and causal 21-step moving averages for RL. Circles denote SFT validation, and dashed diamond curves denote RL validation every 50 steps.}
    \label{fig:two_stage_train}
\end{figure*}

\subsection{Training Settings}
% 中文：基础模型。VRPRM 的 SFT 阶段从 Qwen3-VL-8B-Thinking 初始化，RL 阶段继续训练所得的 SFT 冷启动检查点。我们选择这一基础模型，是为了利用其强大的多模态推理能力支持图像条件下的全局思考和步骤级验证。
\textbf{Base Model.} The SFT stage of VRPRM is initialized from Qwen3-VL-8B-Thinking, and the RL stage continues from the resulting SFT cold-start checkpoint. We choose this base model for its strong multimodal reasoning capability, which supports image-grounded global thinking and step-level verification.

% 中文：SFT 框架与参数高效微调。我们使用 ms-swift 的 lora_llm 模式，仅在语言模型模块中训练 LoRA，视觉塔不进行全量更新。LoRA 的秩为 16、alpha 为 32，采用 bfloat16 和 SDPA，并启用语言模型与视觉塔的梯度检查点以控制显存。
\textbf{SFT Framework and Adaptation.} We train with \texttt{ms-swift} using \texttt{lora\_llm}, which applies LoRA only to the language-model blocks and avoids full vision-tower updates. We use rank 16, alpha 32, bfloat16 precision, SDPA attention, and gradient checkpointing for both the language model and vision tower to control memory.

% 中文：思考格式与上下文。针对 Qwen3-VL-8B-Thinking，我们设置 preserve_thinking=true 并关闭 add_non_thinking_prefix，避免额外前缀改变 <think> 数据分布。多轮样本包含一次全局思考和多个步骤轮次，因此最大上下文长度设置为 12288。
\textbf{Thinking Format and Context.} For Qwen3-VL-8B-Thinking, we set \texttt{preserve\_thinking=true} and \texttt{add\_non\_thinking\_prefix=false}, preserving its native thinking format without introducing a prefix that changes the \texttt{<think>} distribution. Because each sample contains global thinking followed by multiple scoring turns, we set the maximum sequence length to 12,288.

% 中文：优化和损失。模型训练 3 个 epoch，学习率为 \(1.0\times10^{-4}\)，单设备批大小为 1，梯度累积 8 步，共 513 个优化步。ms-swift 的 default 损失仅作用于助手回复；全局思考的 loss_scale 为 0.5，高置信步骤标签为 2.0，不确定步骤设置 loss=false。
\textbf{Optimization and Loss Masking.} We train for three epochs with learning rate \(1.0\times10^{-4}\), per-device batch size 1, and eight gradient-accumulation steps, for a total of 513 optimizer steps. With \texttt{loss\_scale=default} and \texttt{is\_binary\_loss\_scale=false}, causal LM loss is applied only to assistant responses. Global thinking uses loss scale 0.5, confident step labels use 2.0, and uncertain turns set \texttt{loss=false}.

% 中文：RL 阶段。使用 verl（EasyR1 fork）训练：AdamW bf16、学习率 1e-6、权重衰减 1e-2，共 600 个优化步。初始策略为 LoRA 权重合并进基座后的 SFT 检查点。策略损失为 GSPO-token（clip 比率 3e-4 至 4e-4），优势为 GRPO 风格组相对优势；KL 以损失项形式约束到冻结参考策略（低方差 KL，系数 0.2），不使用奖励侧 KL 或在线过滤。每个优化步从 40K 池中无放回采样 32 个 prompt、每个 prompt 16 条 rollout（每步共 512 条生成，全程共 19,200 次提示呈现），温度 0.8、top-p 0.95；guided decoding 强制 <think>/<answer> 结构。RL 使用 LoRA（rank 16、alpha 32，排除视觉模块），视觉塔冻结；每 50 步验证并保存 checkpoint，所有报告结果均使用最终（第 600 步）checkpoint。
\textbf{RL Stage.} We train with \texttt{verl}~\cite{sheng2024hybridflow} (EasyR1 fork) using AdamW in bfloat16 with a learning rate of \(1\times10^{-6}\) and weight decay \(1\times10^{-2}\), for 600 optimization steps. The initial policy is the SFT checkpoint with its LoRA weights merged into the base model. The policy loss is clipped GSPO-token with clip ratios \(3\times10^{-4}\) to \(4\times10^{-4}\), combined with GRPO-style group-relative advantages; the KL term is applied as a loss against the frozen reference policy (low-variance KL, coefficient \(0.2\)), with no reward-side KL and no online filtering. Each optimization step samples 32 prompts without replacement from the 40K pool, with 16 rollouts each (512 generations per step, 19,200 prompt presentations across the run) at temperature 0.8 and top-p 0.95, and guided decoding enforces the \texttt{<think>}/\texttt{<answer>} structure. RL uses LoRA (rank 16, alpha 32, excluding vision modules) with the vision tower frozen, and validates and saves checkpoints every 50 steps; all reported results use the final (step-600) checkpoint. Rollouts follow the constrained format in Appendix~\ref{Appendix:RLReward} and use its combined step, format, and thinking-length reward.

% 中文：continued-SFT 对照（表 2）。从 VRPRM-SFT 检查点（LoRA 合并后）初始化，在同一个 40K 池的无思考逐步序列化版本（共 13,501 条训练对话）上继续监督训练 1 个 epoch（有效 batch 16，共 844 个优化步），沿用 lora_llm 配方与 1e-4 学习率（cosine 调度、5% warmup），最终训练 loss 0.0492。第 500 步验证 loss 0.0676 / token acc 0.9880，训练结束 0.0635 / 0.9894；仅保存并使用最终 checkpoint，与 RL 阶段的选择策略一致。单卡 H200 耗时 5,177 秒（约 1.4 小时），峰值显存 56.5 GiB。
\textbf{Continued-SFT Control.} The control in Table~\ref{Table:PRMBench} initializes from the VRPRM-SFT checkpoint (LoRA merged into the base model) and continues supervised training on the no-thinking stepwise serialization of the same 40K pool, which yields 13,501 training conversations. It runs for one epoch (844 optimizer steps at an effective batch size of 16) with the \texttt{lora\_llm} recipe, learning rate \(1\times10^{-4}\), a cosine schedule, and 5\% warmup, reaching a final training loss of 0.0492. Validation at step 500 gives loss 0.0676 with token accuracy 0.9880, and at the end of training 0.0635 with 0.9894; only the final checkpoint is saved and used, matching the final-checkpoint policy of the RL stage. The run takes 5,177 seconds ($\approx$1.4 hours) on one H200 GPU with a peak memory of 56.5 GiB.

% 中文：启动方式与硬件。本文所有报告的训练与评测均使用 NVIDIA H200 GPU。发布的 SFT 配方默认使用单张可见 GPU，批大小和梯度累积参数按单卡显存设计，也可通过 torchrun 切换为分布式训练。发布的 RL 配方默认在单节点两卡上以 vLLM 托管 rollout 训练，并可通过 verl 的标准多节点启动器扩展。
\textbf{Launch and Hardware.} All training and evaluation runs reported in this paper used NVIDIA H200 GPUs. The released SFT recipe defaults to one visible GPU, with batch size and accumulation chosen for single-device memory, and optionally switches to distributed training through \texttt{torchrun}. The released RL recipe trains with collocated vLLM~\cite{kwon2023vllm} rollouts on two GPUs of a single node by default and scales through the standard \texttt{verl} multi-node launchers.

\subsection{Inference Settings}
% 中文：评估通过兼容 OpenAI API 的 vLLM 服务完成。模型先以温度 0、最多 1024 个词元生成一次全局 <think>，图像只在该阶段输入；随后每个步骤请求复用全局思考和历史判断，以温度 0 和 guided choice 将输出限制为单词元 0 或 1。该方式降低了重复生成开销，并避免步骤得分数量不匹配。无 thinking 变体跳过全局思考请求，由单次包含问题、完整候选解答和相同图像的请求一次性返回全部步骤得分的 JSON 数组；两种协议因此对每个样本恰好暴露一次相同的视觉输入。表 \ref{tab:efficiency_analysis} 的延迟为经该端点在 H200 GPU 上测得的单样本墙上时间：多样本以并发异步请求打分、由服务端持续组批，且每个样本的全局思考只生成一次并在其全部步骤请求间复用。
We evaluate through an OpenAI-compatible vLLM endpoint. The model first generates one global \texttt{<think>} block with temperature 0 and at most 1,024 tokens; images are supplied only at this stage. Each step request then reuses the global thinking and previous judgments, with temperature 0 and guided choice restricting output to the single token 0 or 1. This reduces repeated generation and prevents score-count mismatches. In the no-thinking variant, the global-thinking request is skipped and a single request containing the question, the complete candidate solution, and the same images returns all step scores as one JSON array; both protocols therefore expose the same visual input exactly once per example. Latency in Table~\ref{tab:efficiency_analysis} is per-example wall-clock time measured through this endpoint on H200 GPUs: multiple examples are scored by concurrent asynchronous requests, the server batches them continuously, and the global thinking of each example is generated once and reused across all of its step requests.

\subsection{Metric Conventions on VisualProcessBench}
% 中文：VisualProcessBench 度量约定。五个子集样本量不均衡（MathVerse 1026、MathVision 712、DynaMath 570、WeMath 291、MMMU 267，共 2866 条）。正文 Overall 列遵循官方协议：先池化五个子集的全部步骤，再按正/负两类计算 macro F1，因此大子集权重更高；各子集的无权重均值可直接由子集列推导。真值中性步骤（标签 0）被排除。VisualPRM-8B 使用官方推理代码与官方决策阈值 0.85 复测，与原文数字的微小差异源于本次重测。
The five subsets are uneven in size: MathVerse contributes 1,026 samples, MathVision 712, DynaMath 570, WeMath 291, and MMMU 267 (2,866 in total). The Overall column in the main paper follows the official protocol: all steps across the five subsets are pooled and macro F1 is computed over the correct and incorrect classes, so larger subsets receive proportionally more weight; the unweighted mean of the per-subset values can be derived directly from the per-subset columns. Ground-truth neutral steps (label 0) are excluded. VisualPRM-8B is re-evaluated with its official inference code and its official decision threshold (0.85); minor deviations from the originally reported numbers stem from this re-run.

\section{Two-Stage Training Dynamics}
\label{appendix:train_dynamics}

% 中文：图中同时给出 SFT 与 RL 两阶段的训练动态。SFT 训练损失从 1.006 降至 0.268；验证损失在第 300 步达到最低值 0.3701，同时验证 token accuracy 达到最高值 0.7901。此后的验证曲线仅有小幅波动，表明模型在约第 300 步后进入平台期，而没有出现明显的验证性能崩溃。
Figure~\ref{fig:two_stage_train} reports the optimization dynamics of both stages. During SFT, training loss decreases from 1.006 to 0.268. Validation loss reaches its minimum of 0.3701 at step 300, where validation token accuracy also reaches its maximum of 0.7901. Later validation measurements fluctuate only slightly, indicating a plateau after approximately 300 steps rather than a pronounced validation collapse.

% 中文：完整 RL 日志覆盖第 0 至 600 步，并每 50 步进行一次验证。组采样使批次奖励存在预期波动，但 21 步移动平均在训练过程中保持稳定：前 50 步到最后 50 步的总体奖励、步骤奖励和 macro F1 均值分别从 0.4569、0.4425 和 0.4108 提升至 0.4653、0.4512 和 0.4202。验证总体奖励在第 150 步达到最高值 0.5446（步骤奖励 0.5324，macro F1 0.4975）；最终第 600 步 checkpoint 仍保持 0.5361，差异处于组采样的批次波动范围内，因此所有报告结果均使用该最终 checkpoint，不做基于验证集的挑选。完整训练中格式、思考和数量得分均值分别为 0.9109、0.9412 和 0.6239，表明模型保持了 SFT 学到的输出结构，而主要优化信号仍来自步骤正确性。
The complete RL log covers steps 0--600 with validation every 50 steps. Grouped rollouts introduce expected batch-level variance, while the 21-step moving averages remain stable throughout training: from the first 50 to the final 50 optimization steps, mean overall reward, step reward, and macro F1 increase from 0.4569, 0.4425, and 0.4108 to 0.4653, 0.4512, and 0.4202, respectively. Validation overall reward peaks at 0.5446 at step 150 (step reward 0.5324, macro F1 0.4975); the final step-600 checkpoint retains a validation overall reward of 0.5361, a difference within the expected batch-level variance of grouped rollouts, so all reported results use this final checkpoint without validation-based selection. Across the full run, format, thinking, and count rewards average 0.9109, 0.9412, and 0.6239. These stable auxiliary signals indicate that RL preserves the SFT output structure while optimization remains driven primarily by step correctness.

\section{Example of VRPRM Scoring}
\label{app:scoring_example}
\begin{figure*}[t]
\centering
\begin{tcolorbox}[title=\textbf{Problem Input (Question \& Image)}, width=\textwidth, fonttitle=\bfseries]
    \begin{minipage}{0.65\textwidth}
        \textbf{Question:} Which set does the Figure belong to? Select from A, B, and C.\\
        (A) Set A\\
        (B) Set B\\
        (C) Neither set A nor set B
    \end{minipage}%
    \begin{minipage}{0.3\textwidth}
        \centering
        \includegraphics[width=0.9\linewidth]{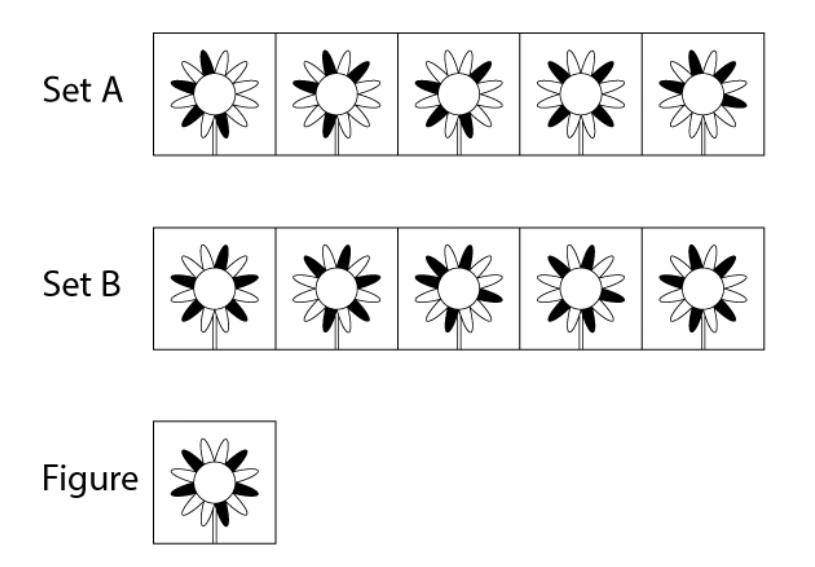}
    \end{minipage}
\end{tcolorbox}
% 中文：VRPRM 评分案例的视觉逻辑问题输入。
\caption{Visual-logic input used in the VRPRM scoring example.}
\label{fig:ExInputScoring}
\end{figure*}

% 中文：为展示 VRPRM 的细粒度验证能力，我们给出 LogicVista 测试集上的案例。该案例不仅对比最终答案，还展示三个关键行为：从示例中归纳视觉规则、定位候选解答中的首个关键错误，以及将错误判断传播到依赖该错误的后续步骤。问题输入见图，正确与错误候选的完整评分见表。
We illustrate fine-grained verification with a LogicVista case study. Beyond contrasting final answers, the example exposes three behaviors: inducing a visual rule from demonstrations, localizing the first consequential error, and propagating the judgment to downstream steps that depend on it. Figure~\ref{fig:ExInputScoring} shows the input, and Table~\ref{tab:vrprm_example_full} compares the complete scoring traces of a correct and an incorrect candidate.

% --- Problem Context Box ---

% --- Detailed Comparison Table ---

\begin{table*}[t]
    \centering
    \scriptsize
    \begin{tabular}{p{0.47\textwidth} | p{0.47\textwidth}}
        \toprule
        \multicolumn{1}{c|}{\textbf{Candidate 1 (Correct)}} & \multicolumn{1}{c}{\textbf{Candidate 2 (Incorrect)}} \\
        \midrule
        \rowcolor{gray!10} \multicolumn{2}{c}{\textbf{1. Policy Generated Step List}} \\

        \textbf{Step 1:} Infer the separating rule from the example flowers in Sets A and B. \newline
        \textbf{Step 2:} Each Set A flower has four shaded petals, whereas each Set B flower has five. \newline
        \textbf{Step 3:} The query flower has five shaded petals. \newline
        \textbf{Step 4:} It therefore follows the rule of Set B. \newline
        \textbf{Step 5:} Answer: $\boxed{B}$
        &
        \textbf{Step 1:} Infer the separating rule from the example flowers in Sets A and B. \newline
        \textbf{Step 2:} Each Set A flower has four shaded petals, whereas each Set B flower has five. \newline
        \textbf{Step 3:} The query flower has four shaded petals. \newline
        \textbf{Step 4:} It therefore follows the rule of Set A. \newline
        \textbf{Step 5:} Answer: $\boxed{A}$
        \\
        \midrule
        \rowcolor{gray!10} \multicolumn{2}{c}{\textbf{2. Global Thinking and Binary Step Judgments}} \\

        \textbf{\textless think \textgreater} The separating rule is the number of shaded petals: Set A has four and Set B has five. The query has five, so the candidate's visual count and conclusion are supported. \textbf{\textless /think \textgreater} \newline
        \textbf{Judgments:} Step 1 $\boxed{1}$; Step 2 $\boxed{1}$; Step 3 $\boxed{1}$; Step 4 $\boxed{1}$; Step 5 $\boxed{1}$.
        &
        \textbf{\textless think \textgreater} The candidate derives the set rule correctly but miscounts the query in Step 3: it has five shaded petals, not four. Steps 4 and 5 depend on this visual error and are therefore invalid. \textbf{\textless /think \textgreater} \newline
        \textbf{Judgments:} Step 1 $\boxed{1}$; Step 2 $\boxed{1}$; Step 3 $\boxed{0}$; Step 4 $\boxed{0}$; Step 5 $\boxed{0}$.
        \\
        \midrule
        \rowcolor{gray!10} \multicolumn{2}{c}{\textbf{3. VRPRM Step Rewards ($P_{\theta}(1 \mid \text{context})$)}} \\

        \begin{tabular}{ll}
            Step 1: & 1.0 \\
            Step 2: & 1.0 \\
            Step 3: & 1.0 \\
            Step 4: & 1.0 \\
            Step 5: & 1.0 \\
            \textbf{Avg:} & \textbf{1.0} (Preferred)
        \end{tabular}
        &
        \begin{tabular}{ll}
            Step 1: & 1.0 \\
            Step 2: & 1.0 \\
            \textbf{Step 3:} & \(\mathbf{5.22{\times}10^{-5}}\) \textcolor{red}{$\downarrow$ First error} \\
            Step 4: & \(1.81{\times}10^{-7}\) \\
            Step 5: & \(3.29{\times}10^{-6}\) \\
            \textbf{Avg:} & \textbf{0.40} (Not Preferred)
        \end{tabular}
        \\
        \bottomrule
    \end{tabular}
    % 中文：视觉逻辑任务上的 VRPRM 评分示例。正确候选的步骤得分均为 1.0；错误候选在首次错误计算阴影花瓣数量的第 3 步得分降至接近零，依赖该视觉错误的后续结论也持续获得低分。
    \caption{\textbf{VRPRM scoring on a visual-logic task.} All steps of the correct candidate receive score 1.0. For the incorrect candidate, the score drops to near zero at Step 3, where the shaded petals are first miscounted, and remains low for the dependent conclusions.}
    \label{tab:vrprm_example_full}
\end{table*}

% 中文：先前资产的许可证与发布条款。VRPRM 构建于下列已核实许可证的先前资产之上：基座模型 Qwen3-VL-8B-Thinking 以 Apache 2.0 发布；训练数据来源 VisualPRM400K-v1.1-Raw 与评测基准 VisualProcessBench 以 MIT 发布；Kimi-K2.6 教师所属的 Kimi K2 系列以 MIT 发布（经商业 API 访问，不再分发模型本身）；训练与推理框架 ms-swift、verl、vLLM 均以 Apache 2.0 发布。上述许可证均允许研究使用及在署名前提下构建衍生作品。
\section{Licenses of Prior Assets and Release Terms}
VRPRM builds on prior artifacts whose licenses we verified before use: the base model Qwen3-VL-8B-Thinking is released under Apache 2.0; the training-data source VisualPRM400K-v1.1-Raw and the evaluation benchmark VisualProcessBench are released under MIT; the Kimi-K2.6 teacher belongs to the Kimi K2 series released under MIT and was accessed through a commercial API without redistributing the model itself; and the training and inference frameworks ms-swift, verl, and vLLM are released under Apache 2.0. All of these licenses permit research use and the construction of derived artifacts with attribution.

% 中文：随论文发布的训练与评测代码将注明上述上游项目，并采用与其条款兼容的许可证分发；任何衍生自 VisualPRM400K-v1.1-Raw 的数据制品保留其来源信息，并依其 MIT 许可证一致的方式发布。评测基准（VisualProcessBench 及其五个子集、表 4 的六个 Best-of-N 基准）仅用于非商业研究评测、不再分发，其创建者在使用处均已引用。
Our released training and evaluation code acknowledges these upstream projects and is distributed under a license compatible with their terms. Any data artifacts derived from VisualPRM400K-v1.1-Raw preserve its provenance and are released consistently with its MIT license. The evaluation benchmarks---VisualProcessBench with its five subsets and the six Best-of-N benchmarks of Table~\ref{Table:BoN}---are used solely for non-commercial research evaluation and are not redistributed; their creators are cited wherever they are used.

% 中文：AI 助手使用披露。写作过程中，作者使用基于大语言模型的 AI 助手完成：（i）将作者撰写的中文草稿翻译为英文并润色英文表达；（ii）LaTeX 排版、文献检索与参考文献信息核对；（iii）训练与评测脚本的实现辅助。全部研究问题、方法设计、实验决策与结果分析均由作者独立完成；所有 AI 辅助产出均经作者审阅、核实与修改，文责由作者承担。实验中作为教师与策略模型使用的大语言模型（Kimi-K2.6、InternVL2.5）及基座模型（Qwen3-VL-8B-Thinking）属于本研究的研究对象，而非写作助手。
\section{Disclosure of AI Assistant Usage}
In preparing this work, the authors used LLM-based AI assistants for (i) translating author-prepared Chinese drafts into English and polishing the English presentation, (ii) LaTeX typesetting, literature search, and verification of bibliographic metadata, and (iii) implementation assistance for the training and evaluation scripts. All research questions, the method design, all experimental decisions, and the analysis of results were conceived and conducted solely by the authors, and every piece of AI-assisted material was reviewed, verified, and edited by the authors, who take full responsibility for the content of this paper. The large language models serving as the teacher and policy models in our experiments (Kimi-K2.6 and InternVL2.5) and the base model (Qwen3-VL-8B-Thinking) are research objects of this study rather than writing assistants.

\end{document}

%% file: Table4BoN.tex
\begin{table*}[t!]
\centering
\scriptsize
\setlength{\tabcolsep}{2.5pt}
\renewcommand{\arraystretch}{1.00}
\setlength{\aboverulesep}{0.25ex}
\setlength{\belowrulesep}{0.35ex}
\setlength{\abovecaptionskip}{3pt}
\begin{tabular}{@{}lccccccc@{}}
\toprule
\textbf{Model} & \textbf{MathVista} & \textbf{MathVision} & \textbf{MathVerse-VO} & \textbf{WeMath} & \textbf{LogicVista} & \textbf{MMMU} & \textbf{Overall} \\
\midrule
\multicolumn{8}{c}{\textbf{Proprietary Models}} \\
\midrule
GPT-4o & 60.00 & 31.20 & 40.60 & 45.80 & 52.80 & 70.70 & 50.18 \\
Gemini-2.0-Flash & 70.40 & 43.60 & 47.80 & 47.40 & 52.30 & 69.90 & 55.23 \\
Claude-3.5-Sonnet & 65.30 & 35.60 & 46.30 & 44.00 & 60.40 & 66.40 & 53.00 \\
\midrule
\multicolumn{8}{c}{\textbf{Open-source Policies and PRM-Guided Selection}} \\
\midrule
InternVL2.5-8B & 58.70 & 18.95 & 26.65 & 30.00 & 35.12 & 52.00 & 36.90 \\
$+$ VisualPRM & 67.50 & 24.44 & 33.88 & 35.81 & 41.16 & 55.33 & 43.02 \\
\quad $\Delta$ over policy & $+8.80$ & $+5.49$ & $+7.23$ & $+5.81$ & $+6.04$ & $+3.33$ & $+6.12$ \\
$+$ \textbf{VRPRM-SFT} & 70.10 & 33.92 & 37.26 & 44.89 & 64.37 & 50.80 & 50.22 \\
\quad $\Delta$ over policy & $+11.40$ & $+14.97$ & $+10.61$ & $+14.89$ & $+29.25$ & $-1.20$ & $+13.32$ \\
$+$ \textbf{VRPRM-RL} & 78.70 & 42.80 & 45.18 & 53.05 & 72.93 & 59.33 & 58.67 \\
\quad $\Delta$ over policy & $+20.00$ & $+23.85$ & $+18.53$ & $+23.05$ & $+37.81$ & $+7.33$ & $+21.77$ \\
\midrule
InternVL2.5-26B & 64.50 & 23.65 & 32.36 & 28.48 & 41.38 & 53.33 & 40.62 \\
$+$ VisualPRM & 71.00 & 27.70 & 39.72 & 40.19 & 47.65 & 56.00 & 47.04 \\
\quad $\Delta$ over policy & $+6.50$ & $+4.05$ & $+7.36$ & $+11.71$ & $+6.27$ & $+2.67$ & $+6.42$ \\
$+$ \textbf{VRPRM-SFT} & 74.70 & 40.23 & 44.81 & 44.84 & 68.78 & 53.94 & 54.55 \\
\quad $\Delta$ over policy & $+10.20$ & $+16.58$ & $+12.45$ & $+16.36$ & $+27.40$ & $+0.61$ & $+13.93$ \\
$+$ \textbf{VRPRM-RL} & \underline{82.40} & 47.50 & \underline{53.17} & 52.67 & \underline{76.73} & \underline{71.78} & \underline{64.04} \\
\quad $\Delta$ over policy & $+17.90$ & $+23.85$ & $+20.81$ & $+24.19$ & $+35.35$ & $+18.45$ & $+23.42$ \\
\midrule
InternVL2.5-38B & 68.20 & 31.48 & 40.23 & 39.50 & 49.89 & 64.00 & 48.88 \\
$+$ VisualPRM & 74.40 & 35.23 & 42.89 & 50.10 & 53.02 & 64.67 & 52.71 \\
\quad $\Delta$ over policy & $+6.20$ & $+3.75$ & $+2.66$ & $+10.60$ & $+3.13$ & $+0.67$ & $+3.83$ \\
$+$ \textbf{VRPRM-SFT} & 75.50 & \underline{49.82} & 48.08 & \underline{60.56} & 73.78 & 69.23 & 62.83 \\
\quad $\Delta$ over policy & $+7.30$ & $+18.34$ & $+7.85$ & $+21.06$ & $+23.89$ & $+5.23$ & $+13.95$ \\
$+$ \textbf{VRPRM-RL} & \textbf{84.30} & \textbf{58.59} & \textbf{57.36} & \textbf{70.29} & \textbf{83.67} & \textbf{77.78} & \textbf{72.00} \\
\quad $\Delta$ over policy & $+16.10$ & $+27.11$ & $+17.13$ & $+30.79$ & $+33.78$ & $+13.78$ & $+23.12$ \\
\bottomrule
\end{tabular}
% 中文：六个多模态推理基准上的 Best-of-8 结果。MathVerse 使用 Vision-Only（VO）划分；Overall 为六项基准的平均分。每个 PRM 下方的增益行给出相对同一 InternVL2.5 基础策略的绝对百分点变化。粗体和下划线分别表示各列最优和次优结果。
\caption{\textbf{Best-of-8 results on six multimodal reasoning benchmarks.} MathVerse uses the Vision-Only (VO) split, and Overall is the mean across all six benchmarks. Each $\Delta$ row reports the absolute change over the corresponding policy. \textbf{Bold} and \underline{underlined} values denote the best and second-best result in each column.}
\label{Table:BoN}
\end{table*}